\documentclass[10pt,journal,compsoc]{IEEEtran}
 
\ifCLASSOPTIONcompsoc
  \usepackage[nocompress]{cite}
\else
  \usepackage{cite}
\fi
 
\usepackage{soul}

\ifCLASSINFOpdf
\else
\fi

\usepackage{times}
\usepackage{epsfig}
\usepackage{graphicx}
\usepackage{amsmath}
\usepackage{amssymb}
\usepackage{enumitem}
\usepackage{array}
\usepackage{float}
\usepackage{algorithm}
\usepackage{algorithmic}
\usepackage{color}
\usepackage{mathtools}  

\newcolumntype{C}[1]{>{\centering\arraybackslash}p{#1}}

\floatname{algorithm}{Algorithm}

\usepackage[pagebackref=true,breaklinks=true,letterpaper=true,colorlinks,bookmarks=false]{hyperref}
\graphicspath{ {Figures/} } 

\newcommand{\rev}[1]{\textcolor{black}{#1}}
\newcommand{\last}[1]{\textcolor{black}{#1}}
\begin{document}
%
 
\title{Infinite Feature Selection:\protect\\ A Graph-based Feature Filtering Approach}

\author{Giorgio~Roffo,
        Simone~Melzi,~\IEEEmembership{Member, IEEE}, Umberto~Castellani,\\
         Alessandro~Vinciarelli,~\IEEEmembership{Member, IEEE} and~ Marco~Cristani,~\IEEEmembership{Member, IEEE}
\IEEEcompsocitemizethanks{\IEEEcompsocthanksitem G. Roffo and A. Vinciarelli are with the School of Computing Science, University of Glasgow, Glasgow, UK.\protect\\
\protect\IEEEcompsocthanksitem M. Cristani, S. Melzi and U. Castellani are with the Department of Computer
Science, University of Verona, Verona, Italy. \protect\\

\protect\IEEEcompsocthanksitem M. Cristani is also affiliate member of the Institute of Cognitive Sciences and Technologies (ISTC), CNR Italy

}
\protect\thanks{Manuscript received August 5, 2019.}}

\markboth{\textbf{IEEE Transactions on Pattern Analysis and Machine Intelligence [PREPRINT 2020]}} {Roffo \MakeLowercase{\textit{et al.}}: Infinite Feature Selection}

\IEEEtitleabstractindextext{%
\begin{abstract}
--
\rev{We propose a filtering feature selection framework that considers subsets of features as paths in a graph, where a node is a feature and an edge indicates pairwise (customizable) relations among features, dealing with relevance and redundancy principles. 
By two different interpretations (exploiting properties of power series of matrices and relying on Markov chains fundamentals) we can evaluate the values of paths (\emph{i.e.}, feature subsets) of arbitrary lengths, eventually go to infinite, from which we dub our framework \emph{Infinite Feature Selection} (Inf-FS). Going to infinite allows to constrain the computational complexity of the selection process, and to rank the features in an elegant way, that is, considering the value of any path (subset) containing a particular feature. We also propose a simple unsupervised strategy to cut the ranking, so providing the subset of features to keep.
In the experiments, we analyze diverse settings with heterogeneous features, for a total of 11 benchmarks, comparing against  18 widely-know comparative approaches. The results show that Inf-FS behaves better in almost any situation, that is, when the number of features to keep are fixed a priori, or when the decision of the subset cardinality is part of the process.}
--
\end{abstract}

}

\maketitle

\IEEEdisplaynontitleabstractindextext

%
\IEEEpeerreviewmaketitle

\ifCLASSOPTIONcompsoc
\IEEEraisesectionheading{\section{Introduction}\label{sec:introduction}}
\else
\section{Introduction}\label{sec:introduction}
\fi

\IEEEPARstart{O}{ver} the last few decades, successful approaches to machine learning problems have been based initially on hand-crafted features (\emph{e.g.}, SIFT and HOG-like ~\cite{Lowe:SIFT,bay2008speeded,al2018spectral,dalal2005histograms}, dictionary-based~\cite{BoVW2003}) that evolved into automatically learned ones with the diffusion of deep learning models~\cite{Hinton:2002,Hinton:2006,bengio2013representation}. 
Through these advancements, feature selection (FS) still remains an active and growing research area that enables both dimensionality reduction and data interpretability, looking for features which are relevant and not redundant~\cite{zeng2011feature,efron2004least,Bradley98featureselection}.

In this paper we introduce a fast graph-based feature filtering approach that ranks and selects features by considering the possible subsets of features as paths on a graph, and works in an unsupervised or supervised setup. 

Our framework is composed by three main steps.
In the first step, an undirected fully-connected weighted graph is built, where the node $\vec v_i$, $1\le i\le n$, corresponds to the feature $f_i$, and each edge connecting $\vec v_i $ to $\vec  v_j$ has associated a weight, or value,  modeling the expectation that features $f_i$ and $f_j$ are relevant and not redundant. The weight comes from customizable pairwise relations among feature distributions, which can be easily crafted by the user, and, as a future perspective, learned directly from data. 
Here we present two instances of pairwise relations: one exploiting class information (Inf-FS$_S$), the other one being completely agnostic (Inf-FS$_U$).

\rev{In the second step, the weighted adjacency matrix \last{associated to the graph} is employed to assess the value of each feature (\emph{i.e.}, a node in the graph) while considering possible subsets of features (\emph{i.e.}, subsets of nodes) as they were paths of variable length.  
Two interpretations can be exploited: one comes from the properties of power series of matrices, the other one from the concept of absorbing Markov chain. In both the cases, we compute a vector which at the $i$-th entry expresses the value (or probability) of having a particular feature in a subset of any length, summing for all the possible lengths, until infinite. Going to infinite allows us to reduce the computational complexity from $\mathcal{O}(n^3lT)$ ($n$ features, $l$ path length, $T$ samples) to $\mathcal{O}(n^3T)$. For this reason, we dubbed our approach \emph{Infinite Feature selection (Inf-FS)}. Ranking the values of the ``infinite'' vector gives the ordered importance of the features.}
\rev{
In the third step, a threshold over the ranking is automatically selected by clustering over the ranked value. The rationale is to individuate at least two distributions, one which contains the features to keep with higher value, the other the ones to discard.}

\rev{The proposed framework is compared against $18$ comparative approaches of feature selection, with the goal of feeding the selected features into an SVM classifier.}  

As for the datasets, we selected 11 publicly available benchmarks to deal with diverse FS scenarios and challenges. In particular we consider five DNA microarray datasets for cancer classification (\emph{Colon}~\cite{alon}, \emph{Lymphoma}~\cite{Golub99}, \emph{Leukemia}~\cite{Golub99}, \emph{Lung}~\cite{Gordon02},  \emph{Prostate}~\cite{citeulike:1624492}), handwritten character recognition (GINA~\cite{GINA}), general classification tasks from the NIPS feature selection challenge (MADELON, GISETTE~\cite{NIPS2003}, DEXTER~\cite{guyon2004result}), and two classic object recognition datasets with convolutional neural networks (CNNs) features (PASCAL VOC 2007~\cite{pascal-voc-2007} and CalTech 101~\cite{FeiFeiFergusPeronaPAMI}).

One of the most interesting aspects shown in the experiments is the \last{flexibility} of Inf-FS, both in its unsupervised and supervised version: independently on the scenario (small-sample+high dimensional, unbalanced classes, severe interclass overlap, noise) Inf-FS overcomes the competitors, and if not, it gives the second or third best performance, promoting itself as all-purpose feature selection strategy. Another important achievement is in the automatic thresholding, which is simple yet effective in deciding which features to keep on \last{any} dataset. Finally, Inf-FS operates also on neural features, acting over cues that have been the state-of-the-art for long time~\cite{Simonyan14c}.

\rev{The proposed framework generalizes the previously published \textit{Infinite Feature Selection} (Inf-FS)~\cite{roffo2017infinite,Roffo:InfFS:2015} presented as an unsupervised filtering approach, explained by algebraic motivations. Here we introduce a supervised counterpart and a strategy to select a subset of features, supported by a novel alternative way to explain the Inf-FS thanks to Markov chains fundamentals.}\\

The rest of the paper is organized as follows: Sec.~\ref{sec:soa} illustrates the related literature, including the comparative approaches we consider in this study. Sec.~\ref{sec:method} introduces our approach showing how the fully-connected graph is built for both the unsupervised and supervised variants. 
Sec.~\ref{sec:markovProcesses} connects the proposed approach to the absorbing Markov chain framework, deriving the subset selection strategy. Extensive experiments are reported in Sec.~\ref{sec:exp}, and, finally, in Sec.~\ref{sec:conc}, conclusions are given and future perspectives are envisaged.

\section{State Of The Art}\label{sec:soa}

Feature selection algorithms are partitioned into three main classes ~\cite{chandrashekar2014survey,Guyon:2003:IVF:944919.944968}: \textit{filters}, \textit{wrappers} and \textit{embedded} methods. \textit{Filter} methods make use of the intrinsic properties of the data (\emph{e.g.}, correlation, variance, locality, information gain, etc.) to evaluate the value of a feature. In contrast, \textit{wrapper} methods assign an importance score to each feature based on the performance of a predictor, which is considered as a black box.
Finally, \textit{embedded} methods include the feature selection process as part of an internal regression model which estimates the relationships among variables. 
\rev{Inf-FS belongs to the filter approaches, since it deals with the sole properties of the data, without relying on a specific predictor. 
}

Within each of the above families of algorithms, FS techniques can be further classified into two sub-categories, \emph{unsupervised} and \emph{supervised}, depending on the use of class-label information in the selection process. \rev{In this paper we present an example of Inf-FS for both the scenarios.}

Most of the feature selection algorithms evaluates an initial feature set, providing a ranking on them as a final output. Subsequently, the ranking is cut by \emph{subset selection} strategies, commonly performed by cross-validation strategies on validation data~\cite{Guyon:2003:IVF:944919.944968}.  

The section overviews the three families of FS methods, separating their unsupervised and supervised versions.

\subsection{Filter methods}

\subsubsection{Unsupervised approaches}

In unsupervised scenarios, filter methods are mainly based on locality preserving principia found by clustering. 
The Laplacian Score (LS)~\cite{HCN05a} evaluates the value of a feature as its tendency to preserve spatial relationships which ensure intra-cluster proximity. Technically, LS constructs a nearest neighbor graph and ranks high those features that are consistent with Gaussian Laplacian matrix~\cite{HCN05a}. 
Similarly, in the multi-cluster feature selection approach  (MCFS)~\cite{Cai:2010}\last{,}  features are selected based on spectral analysis and solving a sparse regression problem, encouraging the formation of compact clusters. Local learning clustering (LLCFS) method~\cite{zeng2011feature} is a kernel learning method that weights features and exploits the weights to regularize the clustering. Uninformative features are left out before the clustering.
\rev{These solutions, included in the experiments, are computationally expensive since rely on clustering. In contrast, our approach is faster since it only uses intrinsic properties of the data.}

\subsubsection{Supervised approaches}

A standard two-class filter method is \emph{Relief} and its multi-class extension \emph{Relief-F}~\cite{liu2008}. In general\last{,}  the strategy evaluates feature value differences between nearest neighbor pairs and scores features according to how well they contribute to the overall class separation. A common criticism of Relief is that it selects redundant subsets, since it is not controlling feature correlation. A solution is given by the minimum Redundancy and Maximum Relevancy (mRMR) algorithm~\cite{peng2005feature}, minimizing the redundancy and maximizing the relevance of the set of features.
 This is obtained by maximizing the joint mutual information (using Parzen Gaussian windows~\cite{suzuki2009mutual}) between the values of a given feature and the membership to a particular class. The mRMR suffers from an expensive computational cost (i.e., $\mathcal{O}(n^2 T^3)$ where $n$ is the number of features and $T$ the number of samples~\cite{peng2005feature,Guyon:2003:IVF:944919.944968,roffo2017ranking}). 
 Another weakness of mRMR comes with the approximation of the mutual information, which is inaccurate when the number of training samples is small~\cite{suzuki2009mutual}. A faster filter approach is the Fisher score~\cite{Quanquanjournals}, which scores the features individually, according to the ratio of inter-class separation and intra-class variance. 

Several algorithms employ mutual information to select the features. The simple method proposed in~\cite{Hutter:02feature} estimates the mutual information between feature distributions and class labels. All the features are evaluated independently, one by one, obtaining a score used to do ranking. The recent Max-Relevance and Max-Independence (MRI)~\cite{wang2017feature} introduces a relevancy additional constraint, by maximizing the classification accuracy while minimizing the redundancy between features. Other  approaches such as CIFE~\cite{lin2006conditional}, MIFS~\cite{battiti1994using} and ICAP~\cite{jakulin2005machine} quantify the redundancy (or dependency) among the set of feature distributions by proposing slightly different variations of the objective function, \emph{i.e.}, the conditional likelihood of the training labels. Similarly, the joint mutual information (JMI)~\cite{yang2000data} and conditional mutual information (CMIM)~\cite{fleuret2004fast} may be included in this group. The common assumption behind all these methods is that independency among features can positively affect the classification performance.

\rev{The Inf-FS framework is attractive because, when computing the weighted adjacency matrix, allows to include inter/intra class reasoning without relying on a specific strategy: in fact, the supervised Inf-FS$_S$ proposed here makes use of a fast computation of the mutual information and the \last{Fisher} criterion, but other alternatives are possible. Especially in the case of large number of samples, mutual information may be dropped in favor of other relations faster to be computed. Another difference with Inf-FS is that the MI-based approaches take into account pairwise (feature-class label) dependencies, while our approach extends the \last{$2$-{nd}} order  to \last{$n$-{th}} order by considering subsets of features as paths on a graph. 
}

Recently, other graph-based approaches have been proposed such as the eigenvector centrality (ECFS) ~\cite{RoffoECML16,Roffo2017b,RoffoBMVC2016} and the infinite latent feature selection (ILFS)~\cite{roffo2017infinite}, which is an extension of the unsupervised Inf-FS$_U$. 
The ECFS ranks features according to a centrality measure over the graph of features (eigenvector centrality), and should be considered a lighter version of Inf-FS$_U$, see  Sec.~\ref{sec:method} for further details. 
\rev{In ILFS, the features are grouped into token by probabilistic latent semantic analysis (PLSA), which in practice learns the weights of the adjacency graph of Inf-FS as to provide better class separability. Instead, our framework requires to explicitly craft the weights; despite the experiments show that our approach overcomes ILFS, we think learning the weights is a convenient direction, which we are interested at the present moment.}

Summarizing, some advantages of using filter methods are: 
\begin{itemize}
    \item faster than wrapper and embedded methods;
    \item scalable;
    \item classifier independent (better generalization).
\end{itemize}
On the other hand, disadvantages are related to a generic lower performance if compared to supervised approaches, since filters are independent on the specific classifier.

\subsection{Wrapper approaches}

\subsubsection{Unsupervised approaches}
In the \last{dependence-guided} unsupervised feature selection (DGUFS)~\cite{JunGuo_AAAI_2018_DGUFS}, 
 feature selection is performed by graph-based clustering through the optimization of two terms: one term individuates dependence among samples by clustering, 
 while the other term votes for features which minimize the intra-cluster variance. 
This approach is showed to be prone to local minima.
The feature selection with adaptive structure learning (FSASL)~\cite{du2015unsupervised} is an iterative approach that captures the global structure of data within a sparse representation framework, where the reconstruction coefficient is learned from the selected features. Its main drawback is the high computational complexity (see Table~\ref{table:compmethods}). 
Finally, the unsupervised feature selection with ordinal locality (UFSOL)~\cite{guo2017unsupervised}  is a clustering-based approach that preserves the relative neighborhood proximities of the samples through distance-based clustering.

\rev{Similarly to our approach, these last three methods estimate inter-relationships among features, but in an iterative fashion which makes the entire process 
expensive and prone to local minima. Conversely, Inf-FS is one-shot.}

\subsubsection{Supervised approaches}
The support vector machine with recursive feature elimination (RFE)~\cite{Guyon:2002} is a popular wrapper method that eliminates useless features in a sequential, backward fashion, ranking high a feature if it separates (by a linear SVM) samples of different classes. 
However, the performance of the RFE becomes unstable at some values of the filter-out factor (i.e., the number of features eliminated in each iteration)~\cite{tang2007development}. To overcome this weakness many different variants of RFE have been proposed, where the initial feature subset is selected using several SVM models with different filter-out factors, and in the second stage, features are selected by eliminating one feature at each iteration. For example, the sample weighting \emph{SW SVM-RFE}~\cite{LeiYi10.1109} gives more weight to those samples that are close to the separating hyperplane. The \emph{Ensemble SVM-RFE}~\cite{LeiYi10.1109} aggregates the results of several SVM-RFE selectors which are applied to randomized training data. 
Finally, the 
recursive cluster elimination (RCE)~\cite{yousef2007recursive} has been introduced to overcome the RFE instability; 
it is a backward elimination algorithm that combines K-means to identify correlated clusters of features. 

Some advantages of wrapper methods are:
\begin{itemize}
    \item exploit the advantages of specific classifiers;
    \item in general, higher classification accuracy than filters.
\end{itemize}
The advantages of wrappers are also disadvantages, since they are suitable for some data only if their associated classifiers are, limiting the overall portability; additionally, wrappers tend to be computational expensive. \last{On the contrary, Inf-FS is classifier agnostic, focusing only on intrinsic properties of data and their labels.} \rev{We omit the RFE-X approaches in the experiments since they have been already shown to be inferior to Inf-FS in \cite{Roffo:InfFS:2015}.}

\subsection{Embedded methods}
Embedded methods include the selection process as part of an internal regression model (\emph{e.g.}, L1, LASSO regularization, decision tree), and the overall ranking process is less prone to overfitting than wrappers. 
\subsubsection{Unsupervised approaches}
An example of unsupervised embedded method is the L$_{2,1}$-norm regularized discriminative feature selection for unsupervised learning (UDFS)~\cite{Yang:2011}. UDFS optimizes an objective function representing a L$_{2,1}$-norm regularized
minimization problem with orthogonal and locality preserving constraints~\cite{chen2016augmented} so that it simultaneously exploits discriminative information and feature correlations. However, such optimization problems are difficult to solve due to the non-smooth objective function and non-convex constraints~\cite{chen2016augmented}.

\subsubsection{Supervised approaches}
In supervised learning scenarios, support vector machines play a role in many embedded approaches. The Feature Selection concaVe method (FSV)~\cite{Bradley98featureselection} generates a separating plane by maximizing the usual margin, minimizing at the same time the number of dimensions (= features) where the plane is defined. Another SVM-based feature selection approach minimizes the \last{0-norm} with (L0) SVMs~\cite{weston2003use}, encouraging sparsity. 
The least square regression (LSR) has also been frequently employed for feature selection. 
The classical embedding approach is the regression by LASSO~\cite{liu2007computational}, 
where feature selection takes place by selecting the variables that have non-zero weighting coefficients. \rev{For classification, LASSO is modified by exploiting a hinge loss (LASSO$_h$), which penalizes linearly with respect to the correct classification labels~\cite{van2008high}. More recently,  \emph{unhinged} losses have shown to be more robust against biased estimates~\cite{van2015learning},  which are a known issue of LASSO (LASSO$_u$).
 In the experiments we consider as comparative approaches both LASSO$_h$ and LASSO$_u$.}

\rev{Another way to avoid bias comes with non-convex optimization strategies, for example with hard-thresholding approaches, which work under the hypotheses of strong restricted convexity/smoothness of the function to be minimized. Recent hard thresholding approaches are GraHTP~\cite{yuan2014gradient,yuan2017gradient} and NHTP~\cite{zhou2019global}, the latter included as comparative approach.
}

Advantages and disadvantages of using embedded methods are similar to those of wrappers (they depend on external techniques), however, they are less prone to over-fitting. \rev{Inf-FS is conceptually different, being a filter which \emph{prepares} the data to a subsequent, independent classification step}.

\section{Our Approach}\label{sec:method}
We propose two different versions of Inf-FS: the unsupervised Inf-FS$_U$ and the supervised Inf-FS$_S$. 
In both the cases, we build upon a weighted undirected fully-connected graph $G = (V, E)$ with node set $V = \{\vec v_1,...,\vec v_n\}$ representing a set of $n$ feature distributions $F = \{f_1,...,f_n\}$,
and edge set $E$ modeling relations among pairs of nodes (\emph{i.e.}, relations among distributions). In the following, the terms \emph{feature} and \emph{feature distribution} will be used interchangeably.   

Let us represent $G$ with its adjacency matrix $A$, where each of its elements $A(i,j)$,  $1\leq i,j \leq n$, models the confidence that features $f_i$ and $f_j$ (the nodes $\vec v_i$ and $ \vec v_j$) are \emph{both} good candidates to be selected, thanks to an associated weight function $\varphi(\cdot,\cdot)$:

\begin{gather}\label{eq:partzero}
  A(i,j) = 	\varphi( \vec v_i, \vec v_j),
\end{gather}
where $\varphi(\cdot,\cdot)$ is a positive, real-valued function defining the \emph{value} of each edge. In the unsupervised version of our approach, referred as Inf-FS$_U$, the function $\varphi_U(\cdot,\cdot)$ is modeled as a function of both the variance and correlation of the features, while in its supervised form (Inf-FS$_S$), the function $\varphi_S(\cdot,\cdot)$ adds the class information using the Fisher criterion and the mutual information. \rev{It is worth noting that other types of functions can be built, with the only constraint that the higher the value of the function, the stronger the preference of selecting both the features.}

\subsection{Graph Building for Inf-FS$_U$}\label{sec:unsupA}

For the unsupervised scenario, $\varphi_U(\cdot,\cdot)$ is a weighted linear combination of two pairwise measures relating the features $f_{i}$ and $f_{j}$, defined as:
\begin{gather}\label{eq:unsupWeight}
  \varphi_U(\vec v_i,\vec v_j) = \alpha E_{ij} + (1-\alpha) \overline{corr}_{ij},
\end{gather}
with $E_{ij}$ indicating the maximal normalized standard deviation over the two distributions, i.e., $E_{ij} = max\left(\sigma_{i},\sigma_{j}\right)$,
where $\sigma_{i}$ is the standard deviation over the samples $\{f_i\}$, normalized to the range $[0,1]$ by the
maximum standard deviation over the set $F$. The second term is the opposite of the correlation  $\overline{corr}_{ij} = 1 - \left|Spearman(f_i,f_j) \right|$, with $Spearman$ indicating Spearman's rank correlation coefficient. The $\alpha$ is a loading coefficient $\in [0,1]$, with its value being  estimated during the experiments by cross validating on the training set for the classification tasks.

\rev{In practice, $\varphi_U(\cdot,\cdot) \in [0,1]$ analyzes two feature distributions, accounting for the maximal feature dispersion (the standard deviation) and how much they are uncorrelated (the Spearman rank correlation coefficient). }

\subsection{Graph Building for Inf-FS$_S$ }\label{sec:supA}

The Inf-FS$_S$ introduces measures which consider class membership information, where we assume to have $G$ classes into play.

The function $\varphi_S(\vec v_i,\vec v_j)$ is formed by three factors: 
the first is the Fisher criterion~\cite{Duda:2000:PC:954544}:
\begin{equation}
	 \tilde{h}_i = \frac{\left | \mu_{i,1} - \mu_{i,2} \right |^2}{ \sigma_{i,1}^2+\sigma_{i,2}^2},
\end{equation}
where $\mu_{i,\textit{g}}$ and $\sigma_{i,\textit{g}}$ are the mean and standard deviation, respectively, assumed by the $i$-th feature when considering the samples of the $\textit{g}$-th class, $1 \le g \le G$.  The multi-class generalization is given by:
\begin{equation}
	h_i = \frac{ \sum_{g=1}^G  (\mu_{i,g} - \hat{\mu_{i}})^2 }{E_{i}^2}
\end{equation}
where $\hat{\mu_{i}}$ and $E_{i}$ denote the mean and standard deviation of the whole data set corresponding to the $f_i$ feature (i.e.,  $E_{i}^2 = \sum_{g=1}^G  (\sigma_{i,g})^2$).  \rev{This is considering intra-class compactness and inter-class separation induced by different features}. The final scores are normalized to have maximum $1$ and minimum $0$. \rev{The closer $h_i$ to 1, the less redundant is the $i$-th feature, since its domain does not overlap with the other ones}.\\

The second factor is the normalized mutual information $m_{i}$ between the features samples of the i-th class and the class label~\cite{estevez2009normalized}: 
\begin{equation}
	m_{i} = \sum_{y \in Y } \sum _{z \in f_i} p(z, y)log \Big( \frac{p(z,y)}{p(z)p(y)} \Big),
\end{equation}
where $Y$ is the set of class labels and $p(\cdot,\cdot)$ stands for the joint probability distribution. Its normalized version is obtained by normalizing over all the $n$ computed values (one for each feature into play). \rev{In practice, $m_i$ measures the amount by which the knowledge provided by the feature vector decreases the
uncertainty about a class, summed over all the classes.}

The third factor is the normalized standard deviation $\sigma_i$ as computed for the unsupervised case.

The three factors are weighted linearly:
\begin{gather}\label{eq:supWeight1}
s_i= h_i \alpha_1 + m_i \alpha_2 + \sigma_i \alpha_3
\end{gather} 
with $1 \leq i,j \leq n$. The parameters $\alpha_k$ are mixing coefficients, $0 \leq \alpha_k \leq 1$,  $\sum_k \alpha_k = 1$, and their values  have been estimated  during the experiments by cross validating on the training set for the classification tasks. 
\rev{Summarizing, the score $s_i$ indicates how much a feature is not redundant (Fisher criterion) and relevant (mutual information, standard deviation) w.r.t. the other classes. } 

Finally, the weights of the adjacency matrix $A$ are obtained by coupling the correspondent $s$ as follows:
\begin{equation}
\varphi_S(\vec v_i,\vec v_j) = A(i,j) = s_i s_j.
\end{equation}

\rev{It is worth noting that the formulation above is just one among the many possible alternatives that computes the value of features $i$ and $j$ taken together. Studying how to estimate this value in an end-to-end fashion would be probably more effective, and is subject of current work.}

\subsection{Feature Ranking Procedure}\label{sec:ranking}
The Inf-FS procedure can be explained in two ways: with the properties  of  power  series  of matrices,  or borrowing from the  concept  of  absorbing  Markov chain. Next, the analysis with the power  series  of matrices is presented, while the  Markov chain view is given at Sec.~\ref{sec:markovProcesses}.\\

Let $\gamma = \{ \vec v_{0}=i,\vec v_{1}, ..., \vec v_{l-1}, \vec  v_{l}=j \}$ denote a path of length $l$ between nodes $i$ and $j$, that is, features $f_i$ and $f_j$, passing through generic nodes $\vec v_{1},...,\vec v_{l-1}$. Let us suppose that the length $l$ of the path is less than the total number of nodes $n$ in the graph. In this case, a path is simply a subset of the features. 

We define the overall weight associated to $\gamma$ as
\begin{gather}\label{eq:two}
\mathcal{\pi}_{\gamma } = \prod_{k=0}^{l-1} A(\vec v_{k},\vec v_{k+1}),
\end{gather}
where $\mathcal{\pi}_{\gamma }$ is actually the value of the path accounting for all the features pairs that belong to it.
There can be more than one path of length $l$ connecting nodes $i$ and $j$. Therefore, we define the set $\mathbb{P}_{i,j}^l$ as containing all the paths of length $l$ between two nodes $i$ and $j$. To estimate the overall contribution of all these paths, we calculate the following sum:
\begin{gather}\label{eq:three}
R_{l}(i,j) =\sum_{\gamma \in \mathbb{P}_{i,j}^l }  \mathcal{\pi}_{\gamma},
\end{gather}
which, following standard matrix algebra, gives:
\begin{gather}
 R_{l} = A^{l}, 
\end{gather}
that is, the power iteration of the adjacency matrix $A$.
\last{$R_{l}$ contains now cycles, and in our feature selection view, this is equivalent to evaluate each feature several times, possibly associated to itself in a self-cycle. This is a side effect that arises with this kind of network, but this possibility holds for all the features, and is taken into account by $R_{l}$.}

We can evaluate the \emph{single feature score} for the feature $x^{(i)}$ at a given path length $l$ as

\begin{gather}\label{eq:four}
c_{l}(i) = \sum_{j \in V} R_{l}(i,j) = \sum_{j \in V} A^{l}(i,j).
\end{gather}
In practice, Eq.\ref{eq:four} models the value of the feature $x^{(i)}$ when considered in whatever selection of $l$ features; the higher $c_{l}(i)$, the better.
Therefore, a possible strategy could be that of ordering the features decreasingly by $c_{l}$, taking the first $m$ obtain a relevant set. Unfortunately, the computation of $c_{l}$ is expensive, bounded by ($\mathcal{O}((l-1)\cdot n^3)$): in fact, $l$ is of the same order of $n$, so the computation turns out to be $\mathcal{O}(n^4)$ and becomes impractical for large sets of features to select ($>10K$); our approach addresses this issue 1) by expanding the path length to infinity $l\rightarrow \infty$ and 2) using notions from algebra to analytically solve the ranking problem in a computationally convenient way.

Eq.\ref{eq:four} estimates the score for feature $f_i$ when injected in whatever subset of $l$ features. Taking into account all the possible path lengths can be elegantly modeled by letting $l \rightarrow \infty$.

\begin{gather}\label{eq:fourPOINTfive}
c(i) = \sum_{l=1}^\infty c_{l} (i) =\sum_{l=1}^\infty \big( \sum_{j \in V} R_{l}(i,j) \big).
\end{gather}
Let $C$ be the geometric series of adjacency matrix $A$:
\begin{gather}\label{eq:fourPOINTfive2}
 C = \sum _{l=1}^{\infty} A^{l},
\end{gather}
It is worth noting that $C$ can be used to obtain $c(i)$ as
\begin{gather}\label{eq:preliminaries2}
c(i) = \sum _{l=1}^{\infty} c_{l}(i)=[(\sum_{l=1}^{\infty}A^{l})\textbf{e}]_{i}=[C \textbf{e}]_{i},
\end{gather}
where $\textbf{e}$ indicates a $1D$ vector of ones, and the square bracket indicates the extraction of an entry of the vector, specified by the index $i$.

The problem is, summing infinite $A^l$ terms could lead to divergence; in which case, regularization is needed, in the form of generating functions~\cite{Graham:1994}, usually employed to assign a consistent value for the sum of a possibly divergent series. There are different forms of generating functions~\cite{Bergshoeff}. We define the generating function for the $l$-path as
\begin{gather}\label{eq:five}
\check{c}(i) = \sum_{l=1}^\infty  r^{l} c_{l} (i) = \sum_{l=1}^\infty\sum_{j \in V} r^{l} R_{l}(i,j),
\end{gather}
where $\textit{r}$ is a real-valued regularization factor, and $\textit{r}^{l}$ can be interpreted as the weight for paths of length $l$. The parameter $\textit{r}$ has been defined as  $r=0.9/\rho(A)$, with  $\rho(A)$ spectral radius of A (more on this at Sec.~\ref{sec:regularization}), ensuring that the infinite sum converges. 

From an algebraic point of view, $\check{c}(i)$ can be efficiently computed by using the convergence property of the geometric power series of a matrix (for a proof, see Sec.~\ref{sec:regularization}):
\begin{equation}\label{eq:Ctilde}
          \check{C} = (\textbf{I} - \textit{r} A)^{-1} - \textbf{I},
\end{equation}

Matrix $\check{C}$ encodes the partial scores of our set of features. The goodness of this measure is strongly related to the choice of parameters that define the adjacency matrix $A$.

We can obtain final relevancy scores for each feature by marginalizing this quantity:
\begin{gather}\label{eq:seven}
          \check{c}(i) = [\check{C} \textbf{e}]_{i}.
\end{gather}
Ranking in decreasing order the $\check{c}$ vector gives the output of the algorithm: a ranked list of features where the most discriminative and relevant features are positioned at the top of the list. The gist of the Inf-FS is to provide a score of importance for each feature as a function of the importance of its neighbors. See Algorithms \ref{algorithm1} (unsupervised) and \ref{algorithm2} (supervised) for a sketch of our approaches.

\subsection{Choice of the regularization parameter $r$}\label{sec:regularization}
In this section, we want to justify the correctness of the method in terms of convergence. The value of $r$ (used in the generating function, and introduced in the previous section, Eq.~\ref{eq:five}) can be determined by relying on linear algebra~\cite{HubHub01}. Let us define  $\left\lbrace \lambda _{0},..., \lambda _{n-1} \right\rbrace $ as the eigenvalues of the matrix A; drawing from linear algebra, we can define the spectral radius $\rho (A)$ as:
$$\rho(A) = \max_{\lambda_{i} \in \left\lbrace \lambda _{0},..., \lambda _{n-1} \right\rbrace} \Big( \vert \lambda _{i} \vert \Big) .$$
For the theory of convergence of the geometric series of matrices, we also have::
$$ \lim_{l \to \infty} A ^{l} = 0 \ \Longleftrightarrow \ \rho(A)<1 \ \Longleftrightarrow \  \sum_{l=1}^{\infty} A ^{l} = (\textbf{I} - A)^{-1} - \textbf{I} .$$
Furthermore, Gelfand's formula ~\cite{powers2015mathematical} states that for every matrix norm, we have:
$$ \rho (A) = \lim _{k\longrightarrow \infty } \vert \vert A^{k} \vert \vert ^{\frac{1}{k}}. $$
This formula leads directly to an upper bound for the spectral radius of the product of two matrices that commutes, given by the product of the individual spectral radii of the two matrices, that is, for each pair of matrices A and $B$, we have:
$$ \rho ( AB ) \leq \rho (A) \rho (B) . $$
Starting from the definition of $ \check{s}(i) $ and from the following trivial consideration:
\begin{center}
$r^{l}A^{l} = \left( r^{l}\textbf{I} \right) A^{l} = \left[ \left(r\textbf{I}\right) A \right] ^{l}$,
\end{center}
we can use Gelfand`s formula on the matrices $r\textbf{I}$ and A and thus obtain:
\begin{equation}
\rho \Big( \left( r\textbf{I} \right) A \Big) \leq \rho (r\textbf{I}) \rho (A) = r \rho (A).
\end{equation}
For the property of the spectral radius:
$\lim_{l \to \infty}\left( rA \right) ^{l} = 0 \Longleftrightarrow \rho (rA) < 1 $.
Thus, we can choose $r$, such as $0 <r<\dfrac{1}{\rho(A)}$; in this way we have:
\begin{eqnarray}
0 <\rho (rA) &=& \rho \Big( \left( r\textbf{I} \right) A\Big) \leq \rho (r\textbf{I}) \rho (A)\nonumber\\
 &=& r \rho (A) <\dfrac{1}{\rho(A)} \rho (A) = 1 \,
\end{eqnarray}
that implies $\rho (rA) < 1 $, and so:
$$ \check{C} = \sum _{l=1}^{\infty} (rA)^{l} = (\textbf{I} - \textit{r}A)^{-1} - \textbf{I}$$
This choice of $r$ allows us to have convergence in the sum that defines $\check{c}(i)$. Particularly, in the experiments, we use $r=\dfrac{0.9}{\rho(A)}$, leaving it fixed for all the experiments.

\begin{algorithm}[t!]
\caption{Unsupervised Infinite Feature Selection}
\label{algorithm1}
\begin{algorithmic}
\REQUIRE{$F = \{ \vec f_1, ..., \vec f_n  \}$ , \textbf{$\alpha$}}\\
\ENSURE{$\check{c}$ final scores for each feature }\\
\textbf{+ Building the graph} \\
\FOR{$i = 1 : n$}
\FOR{$j = 1 : n$}
\STATE $\sigma_{ij}=max( std(f_i) ,std(f_j) )$\;\\
\STATE $\overline{corr}_{ij}= 1-|Spearman( f_i,f_j)|$\;\\
\STATE $A(i,j) = \alpha \sigma_{ij} + (1 - \alpha ) \overline{corr}_{ij} $
\ENDFOR
\ENDFOR \\
\textbf{+ Letting paths tend to infinite} \\
\STATE $r =\frac{ 0.9}{\rho (A)}$\\
\STATE $\check{C} = (\textbf{I} - rA)^{-1} - \textbf{I}$\\
\STATE $\check{c} = \check{C} \thinspace \textbf{e}$
\RETURN $\check{c}$
\end{algorithmic}
\end{algorithm}

\begin{algorithm}[t!]
\caption{Supervised Infinite Feature Selection}
\label{algorithm2}
\begin{algorithmic}
\REQUIRE{$F = \{ f_1, ..., f_n  \}$ , $Y = \{ 1, ..., G \}$ , \textbf{$\alpha_1,\alpha_2,\alpha_3$}}\\
\ENSURE{$\check{c}$ final scores for each feature  }\\
\textbf{+ Building the graph} \\
\FOR{$i = 1 : n$}
\STATE	$h_i \frac{ \sum_{k=1}^K  (\mu_{i,k} - \hat{\mu_{i}})^2 }{E_{i}^2}$\;\\
\STATE	$m(i) = \sum_{y \in Y } \sum _{z \in f_i} p(z, y)log \Big( \frac{p(z,y)}{p(z)p(y)} \Big)$
\STATE Compute $\sigma_{i}$\;\\
\STATE $s_i= h_i \alpha_1 + m_i \alpha_2 + \sigma_i \alpha_3$
\ENDFOR
\FOR{$i = 1 : n$}
\FOR{$j = 1 : n$}
\STATE $A(i,j) =  s_i s_j$
\ENDFOR
\ENDFOR\\
\textbf{+ Letting paths tend to infinite} \\
\STATE $r =\frac{ 0.9}{\rho (A)}$\\
\STATE $\check{C} = (\textbf{I} - rA)^{-1} - \textbf{I}$\\
\STATE $\check{c} = \check{C} \thinspace \textbf{e}$
\RETURN $\check{c}$
\end{algorithmic}
\end{algorithm}

\subsection{\rev{An alternative view of Inf-FS as absorbing random walks}}\label{sec:markovProcesses}

This section provides a different perspective of the proposed framework in terms of absorbing Markov chains and random walks.

Following standard theory on stochastic processes~\cite{kemeny1976markov}, any $m \times m$ transition matrix $T$ of a discrete time, first-order Markov chain with $m$ states can be written in the \textit{canonical} form, which separates \emph{absorbing states} (having probability of self-transition $= 1$) from transient ones by re-ordering rows and columns as follows:
\begin{gather}
T =
  \begin{bmatrix}
    \textbf{I} & \textbf{0}  \\
    R & \tilde{A}
  \end{bmatrix}
\end{gather} 
where $\tilde{A}$ is the square submatrix of size $n \times n$ giving the transition probabilities from non-absorbing to non-absorbing states ($n \le m$), $R$ is the non-null rectangular submatrix of size $n \times k$ giving transition probabilities from non-absorbing to absorbing states ($k=m-n$), $\textbf{I}$ is the identity matrix of size $k \times k$, and \textbf{0} is a rectangular matrix of zeros of size $k \times n$.

\last{When $k>0$, it means we have non-null probability of ending in a absorbing state, with  $R$ and $\tilde{A}$ that are both substochastic, meaning that summing (separately) over their rows gives at least one row less than 1; }in the case of $k=0$ we have that the matrices $R,\mathbf{I},\mathbf{0}$ vanish, and the transition matrix $T=\tilde{A}$ \last{is stochastic and} has no absorbing state. In the following, \last{we assume that all of the rows of $\tilde{A}$ are substochastic, so that necessarily there is at least one absorbing state, so that $k>0$}.

With the canonical form, it becomes easy to compute different  quantities, all related to the probability of having a particular random walk associated to $T$. 
In particular, the probability of having a walk of $l$ steps\footnote{Here step means a single iteration of the stochastic process modeled by the Markov chain} from state $i$ to state $j$, $1 \leq i,j \leq m$ is given by  
\begin{equation}
T^l =
\begin{bmatrix}
    \textbf{I} & \textbf{0}  \\
  (\textbf{I}+\tilde{A}+\tilde{A}^2+...+\tilde{A}^{l-1})R & \tilde{A}^l
  \end{bmatrix}\label{eq:before_inf}
\end{equation}
The fact that $\tilde{A}$ is substochastic in all its rows is a sufficient condition which tells us that its spectral radius is $\rho(\tilde{A})<1$ \cite{horn2012matrix}, which is the same condition that we required for the convergence of the infinite sum at Sec.~\ref{sec:regularization}, this implying $\tilde{A} = rA$.
\last{Therefore, let us suppose that $\tilde{A} = rA$, for a specific $r$ which will be discussed next, and $A$ built as described in Sec.~\ref{sec:supA} and Sec.~\ref{sec:unsupA}, so that $\tilde{A}(i,j)$ indicates
the probability of choosing feature $j$ after having selected $i$. Under this probabilistic view, the higher $\tilde{A}(i,j)$, the higher the complimentarity between $j$ and $i$.  Going from a (transient) state of $\tilde{A}$ into an absorbing state $b$, $1\leq b \leq k$, driven by probability $\tilde{A}(i,b)$, would mean to end the feature selection process. Intuitively, a high $\tilde{A}(i,b)$  would mean that no other transient state (feature) $j$, $k+1\leq j \leq k+n$  is complimentary w.r.t. $i$. Following this perspective, we may compute $T^\infty$ as containing the probability of going from two states in an infinite number of steps by rewriting Eq.~\ref{eq:before_inf} with $\tilde{A}\to \infty=\mathbf{0}$ and}
\begin{gather}
T^{\infty} =
\begin{bmatrix}
    \textbf{I} & \textbf{0}  \\
  CR & \textbf{0}
  \end{bmatrix}
\end{gather}
where the matrix
\begin{equation}\label{eq:C}
C = \textbf{I} + \tilde{A} + \tilde{A}^2 + ... + \tilde{A}^{\infty}= (\textbf{I}-\tilde{A})^{-1}
\end{equation}
At this point, interesting facts do emerge:
\begin{itemize}
\item the matrix $C$ of Eq.~\ref{eq:C} resembles the matrix of Eq.~\ref{eq:Ctilde}  $\check{C} = (\textbf{I} - \textit{r} A)^{-1} - \textbf{I}$, with $\tilde{A} = rA$ and  a difference given by the identity matrix $I$. 
\item \last{In the Markov chain hypothesis, matrix $C$ expresses with $C(i,j)$ the expected number of visits to transient state $j$ starting from transient state $i$, before to go into an absorbing state. In our feature selection case, $C(i,j)$ could be seen as the length of the path enabled by feature $i$ before to end the process of selection: a long path means that there is a pool of features, including necessarily $i$ and $j$, which are strongly complimentary among each other (that have high probability to have transitions among themselves). In the same way, considering $c=C\mathbf{e}$, $c_i$ indicates how much, in general, feature $i$ enable long paths, irrespective of the arrival feature $j$. The longer the path, the more complimentary is the feature $i$ with respect to all the other features.}
\item \last{Unfortunately, the matrix $A$ that we build with the procedures in Sec.~\ref{sec:unsupA} and Sec.\ref{sec:supA}, in general, could be not substochastic, neither could be their regularized versions $rA$ of Sec.~\ref{sec:regularization}.} In fact, Sec.~\ref{sec:regularization} indicates a necessary and sufficient condition for making $rA$ convergent to 0 at infinity, which is not sufficient for being substochastic. 
\end{itemize}
The three observations above suggest a different, stronger regularization than the one expressed by Sec.~\ref{sec:regularization} ($r=0.9/\rho(A)$), in order to be compatible with the Markov chain paradigm; in practice, we need to have $rA$ with $r=0.9/r_{max}$, where $r_{max} = \max_i\sum_{j=1}^n A(i,j)$ is the max summation over the rows of the original matrix adjacency $A$. This makes $rA$ both convergent to 0 at infinity, and substochastic, unlocking an alternative, more interpretable view of our selection process. 

At the same time, with the above regularization, the $\check{C}$ of matrix Eq.~\ref{eq:Ctilde} measuring the value of a couple of features at infinity can be computed as the $C$ matrix at Eq.~\ref{eq:C}, and, consequently, the vectors to be ordered become $\check{c}=\check{C}\mathbf{e}$ and $c=C\mathbf{e}$.

\last{It is worth noting that $\check{c}$ and $c$
give rise to the same ranking, so choosing one regularization $r=0.9/\rho(A)$ or the other  $r=0.9/r_{max}$, in practice, makes absolutely no difference: the two regularizations give just two different interpretations of the same process.}


\subsection{\rev{Selection of the number of features}}\label{sec:feature selection}
\last{The vector $\check{c}$ obtained by Eq.\ref{eq:seven} contains at the $i$-th entry, in term of power series of matrices, the cumulative cost of having a particular feature in any (possibly infinite) subset of features. Equivalently, in terms of Markov chain, $c_i$ of Sec.~\ref{sec:markovProcesses} represents the expected number of selections of features which are complimentary to $i$ that have been chosen before to finish the process of feature selection.} 

\last{Ranking the $c$ vector for feature filtering under the former perspective amounts to rank features which ensures paths of higher costs, where the cost, by construction, is higher for features which are relevant and redundant. Choosing the high-ranked features ensures to consider features of high value.  In the Markov chain assumption, ranking the $c$ vector amounts to promote features which are highly complimentary to each other.}

Looking at how the values of $\check{c}$ (or, equivalently, $c$ ) are distributed will give a global view of the features into play. Experimentally, we have found that the features are bipartite (especially in the supervised case), expressing features which are useful for the classification process and features that carry few or no value. In other words, it is easy to spot a structure in this data, which can be extracted by a clustering procedure.

In this paper we propose to select a particular number of features, by  considering  the distribution of the $\{c_i\}$ values, and select by  a clustering method the features which include the first ranked feature. Different clustering strategies can be taken into account: in our case, we consider 1D Mean-shift with automatic bandwidth selection~\cite{comaniciu2001variable}, which showed to be highly effective in the experiments.

Future work will be devoted in looking for alternative ways to cluster the data: in particular, we spot few cases in which the Mean-shift was not working, due to \last{ Pareto-like distributions}.

\begin{table}[!t]
\small
\centering
\resizebox{0.48\textwidth}{!}{%
\begin{tabular}{|C{2.1cm} |C{0.6cm}| C{0.8cm} |C{2.9cm}|}
\hline
\textbf{Acronym} &   \textbf{\small{Type}} & \textbf{\small{Class}} & \textbf{Comp. complexity} \\\hline
 LLCFS \cite{zeng2011feature}  &f&u& N/A  \\\hline
 LS \cite{HCN05a} &f&u& N/A  \\\hline
 MCFS \cite{Cai:2010}&f&u&  N/A  \\\hline
 Relief-F~\cite{liu2008} &f&s& $\mathcal{O}(iTnG)$  \\\hline
 MI~\cite{Hutter:02feature} &f& s&\small{$\mathcal{O}(T^2n^2)$}\\\hline
 Fisher~\cite{Quanquanjournals}   &f&s& $\mathcal{O}(Tn)$   \\\hline
 ECFS ~\cite{RoffoECML16,Roffo2017b} &f&s&  $\mathcal{O}(Tn + n^2)$  \\\hline
 ILFS \cite{roffo2017infinite} &f&s& $\mathcal{O}(n^{2.37}+in+T+G)$  \\\hline
 CFS \cite{Guyon:2002} &f&u& $\mathcal{O}(\frac{n^2}{2}T)$  \\\hline
 UDFS \cite{Yang:2011}  &f&u& N/A \\\hline
 DGUFS\cite{JunGuo_AAAI_2018_DGUFS} & w & u & N/A \\\hline
 FSASL \cite{du2015unsupervised} & w & u & $\mathcal{O}(n^3 + Tn^2)$ \\\hline
 UFSOL\cite{guo2017unsupervised} & w & u & $\mathcal{O}(iTGn^3)$ \\\hline
 RFE \cite{Guyon:2002} & w& s& \small{$\mathcal{O}(T^2 n log_2n )$} \\\hline
 FSV~\cite{Bradley98featureselection} & e& s& \small{$\mathcal{O}(T^2n^2)$} \\\hline
 \rev{LASSO (hinged)~\cite{van2008high}} \rev{(unhinged)~\cite{van2015learning}}& e & s &\small{$\mathcal{O}(T^2n^2)$} \\\hline
 \rev{NHTP~\cite{zhou2019global}}& e & s& N/A \\\hline
 \textbf{Inf-FS$_U$}  &f&u& \last{$\mathcal{O}(n^{3}(1+T))$} \\\hline
 \textbf{Inf-FS}$_S$ &f&s& \last{$\mathcal{O}(T^2 + n^{3}(1+T))$}  \\\hline
\end{tabular}}
\caption{Feature selection approaches considered in the experiments of Sec.~\ref{sec:exp}. The methods follow the taxonomy of Sec.~\ref{sec:soa}, and are characterized by \emph{type} (f=filter, w=wrapper, e=embedded), \emph{class} (u = unsupervised, s = supervised) and \emph{computational complexity}.
As for the complexity, $T$ is the number of samples, $n$ is the number of initial features, $i$ is the number of iterations in the case of iterative algorithms, and $G$ is the number of classes.}
\label{table:compmethods}
\end{table}

\begin{table*}[!]
\small
\centering
\resizebox{0.90\textwidth}{!}{%
\begin{tabular}{l c c c c c c c c c }
\hline \hline
\textbf{Dataset} & \textbf{Ref.} & \textbf{\#Samples} & \textbf{\#Classes} & \textbf{\#Feat.}  & \textbf{\emph{few train}}& \textbf{\emph{unbal. (+/-)}} & \textbf{\emph{overlap}} & \textbf{\emph{noise}} & \textbf{\emph{sparse}}\\\hline
\emph{COLON} &~\cite{alon} & 62 &2& 2K   &  X & (40/22) & \emph{n.s.}& X  &  \\
\emph{LEUKEMIA} &~\cite{Golub99} & 72 &2& 7129   &  X & (47/25) &\emph{n.s.} & X & \\
\emph{LUNG} &~\cite{Gordon02} & 181 &2& 12533   &  X & (31/150) & \emph{n.s.}& X & \\
\emph{LYMPHOMA} &~\cite{Golub99} & 45 &2& 4026   &  X & (23/22) & \emph{n.s.}& &  \\
\emph{PROSTATE} &~\cite{citeulike:1624492}  & 102 & 2 & 6033   &  X & (50/52)&\emph{n.s.}& &  \\
\hline 
DEXTER &~\cite{guyon2004result} &  2600 &2& 20K   &   & (1,3K/1,3K) & X & X  & X\\
GISETTE &~\cite{NIPS2003} &  6000 &2& 5K   &   & (3K/3K) & X & X &  \\
GINA &~\cite{GINA} & 3153 &2& 970   &   & (1,5K/1,6K) & X & & \\  
MADELON &~\cite{NIPS2003} &  2000 &2& 500   &   & (1K/1K) & X & X & \\
\hline \vspace{0.02cm}
VOC 2007 &~\cite{pascal-voc-2007} & ~10K &20& 4096   & & X & X & X & \\
CalTech 101 &~\cite{FeiFeiFergusPeronaPAMI} & ~10K &102& 4096   & & X & X & X & \\
\hline
\end{tabular}}
\caption{Datasets and the challenges for the feature selection scenario. The abbreviation \emph{n.s.} stands for \emph{not specified} (for example, in the DNA microarray datasets, no information on class overlap is given in advance).}
\label{table:ch5datasets}
\end{table*}

\section{Experiments and Results}\label{sec:exp}

In this section, we compare our framework with several feature selection methods considering both recent approaches \cite{zhou2019global,van2015learning,guo2017unsupervised, JunGuo_AAAI_2018_DGUFS,roffo2017infinite,RoffoECML16,Roffo2017b}, as so as some established algorithms \cite{Cai:2010,liu2008,Hutter:02feature,Quanquanjournals,Guyon:2002,Yang:2011,Bradley98featureselection,van2008high}. Methods are selected to cover the three different families presented in Sec.~\ref{sec:soa}, i.e., filter, wrapper and embedded approaches. Tab. \ref{table:compmethods} lists the methods included in the experiments, reporting their \emph{type} (\emph{f} = filters, \emph{w} = wrappers, \emph{e} = embedded methods), and their \emph{class} ( \emph{s} = supervised or \emph{u} = unsupervised). Additionally, the table shows the computational complexity whereas it has been provided.

\begin{figure*}[!]
\centering
\includegraphics[width=0.45\textwidth]{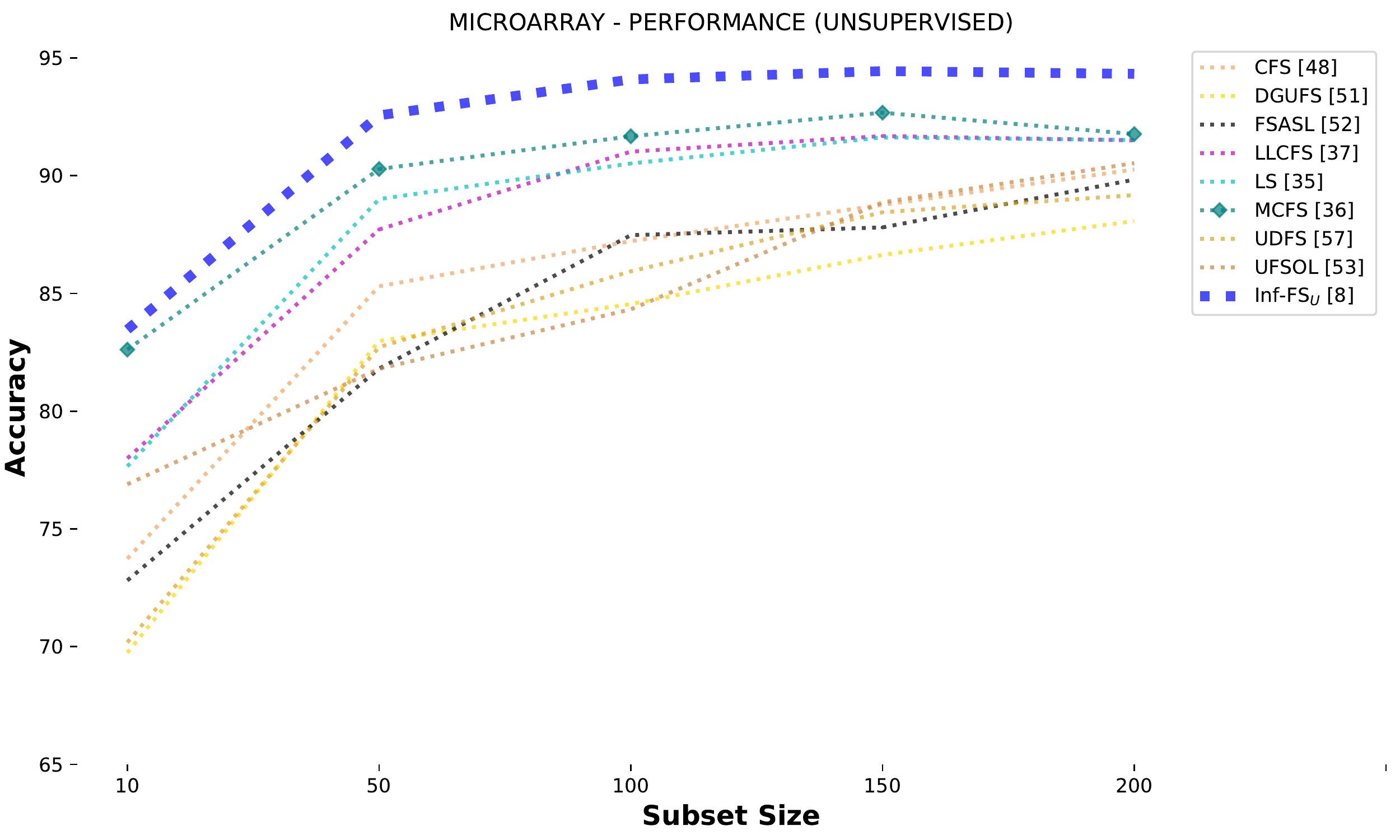}
\includegraphics[width=0.45\textwidth]{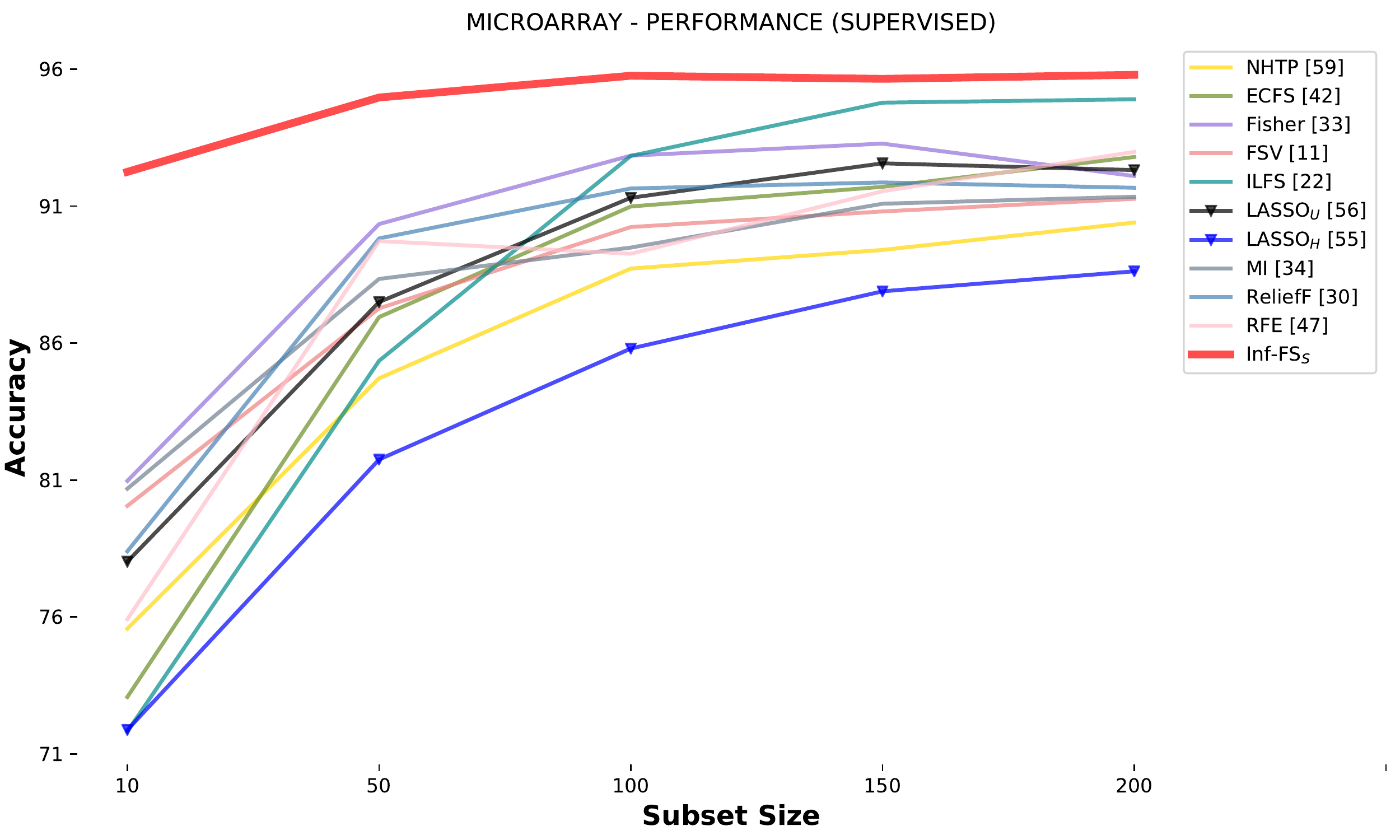}
\caption{Classification results on the small-sample, high-dimensional challenge. On the left, the average performance curves for unsupervised approaches, and on the right, supervised methods are shown. In all of the cases, the performance is measured at different numbers of selected features (on the x-axis). 
}\label{fig:EXPSET1_BIO}
\end{figure*}

The experiments are performed on $11$ different publicly available benchmarks, whose characteristics are summarized in Table~\ref{table:ch5datasets}. The benchmarks allow to evaluate the proposed approach on supervised classification problems, focusing first on small-sample, high-dimensional scenarios, studying the strengths and weaknesses of the unsupervised and supervised Inf-FS on heterogeneous datasets, dealing then with features produced by deep learning algorithms. All of these experiments evaluate the feature selection approaches when they are constrained to provide a definite number $b$ of features; different $b$'s are considered (see in the following sections).
In addition,  we evaluate the automatic subset selection capability, where the optimal number of features has also to be decided.
\rev{A conclusive statistics shows the Inf-FS framework as the most versatile and effective general-purpose algorithm among the considered competitors.}  All of the (MATLAB) code is available at \url{https://github.com/giorgioroffo/Infinite-Feature-Selection}.

\subsection{\emph{Challenge 1}: Small-sample, high-dimensional }\label{sec:BIO}

Treating few samples described by many features is a traditional feature selection challenge. For example, in the medical field~\cite{dernoncourt2014analysis} observations are often difficult to collect (e.g., in the case of rare diseases), while the number of measurements performed on each sample can easily reach the order of thousands (e.g., set of DNA sequences). The small-sample, high-dimensional scenario holds in many other fields like business intelligence~\cite{duan2012business}, geoscience~\cite{pal2010feature} and the automatic analysis of behavioural cues and social signals~\cite{vinciarelli2009social,Scibelli2018}).

Here we consider five widely used small-sample, high-dimensional 2-class microarray datasets: \emph{Colon}~\cite{alon}, \emph{Lymphoma}~\cite{Golub99}, \emph{Leukemia}~\cite{Golub99}, \emph{Lung}~\cite{Gordon02}, and \emph{Prostate}~\cite{citeulike:1624492}. They have been chosen for their variability in terms of number of features (from 2000 to 12533, see Tab.~\ref{table:ch5datasets}) which characterize 45 to 181 samples, because they deal with balanced and unbalanced classes, and because they are widely used in the literature. An exhaustive list of microarray small-sample, high-dimensional datasets can be found in \textit{https://bit.ly/2OSlOfv}, while an essay on generic microarray datasets can be found in~\cite{bolon2014review}.

\rev{The experimental protocol consists in splitting the samples of the dataset in 70\% for training and 30\% for testing. The training procedure consists in building the matrix $A$ as described in Sections \ref{sec:unsupA} and \ref{sec:supA}. In the case of Inf-FS$_S$, the class labels are taken into account, while in the unsupervised case they are ignored. After the training, a selection of the ranked features is considered, by keeping the top-$b$ features, with $b$ variable. The selected features are used to train a linear SVM, where a 5-fold cross-validation on training data is used to set the best $C$ regularization parameter. The same experimental protocol has been applied to all the comparative feature selection approaches}.

The number $b$ of selected features varies (i.e., $b=$10, 50, 100, 150, and 200) in order to show the performance at different regimes. The performance is specified in terms of classification accuracy. In order to avoid any bias induced by a particularly favourable split, this procedure is repeated 20 times by shuffling the data (keeping training and testing separated) and the results are averaged over the trials. A cross-validation is carried out on each training partition of the datasets to select the $\{\alpha\}$ parameters introduced in Sec. \ref{sec:unsupA} and \ref{sec:supA}.

Fig.~\ref{fig:EXPSET1_BIO} depicts the results: on the \emph{left}, the average performance obtained over all of the datasets by the unsupervised approaches are reported; on the \emph{right}, supervised approaches are shown.

On Fig.~\ref{fig:EXPSET1_BIO} (left and right), it can be seen that in both the unsupervised and supervised case, the performance improves substantially with the number of the selected features up to a knee around 50 features; after 150 features, in general, the performance tends to saturate. On the left, it can be seen that Inf-FS$_U$ outperforms the existing methods with a mild but consistent average gap. On the right, Inf-FS$_S$ achieves definitely the best performance, in particular when the number of selected features is fixed to be small (from 10 to 100).

\begin{figure*}[!]
\centering
\includegraphics[width=0.45\textwidth]{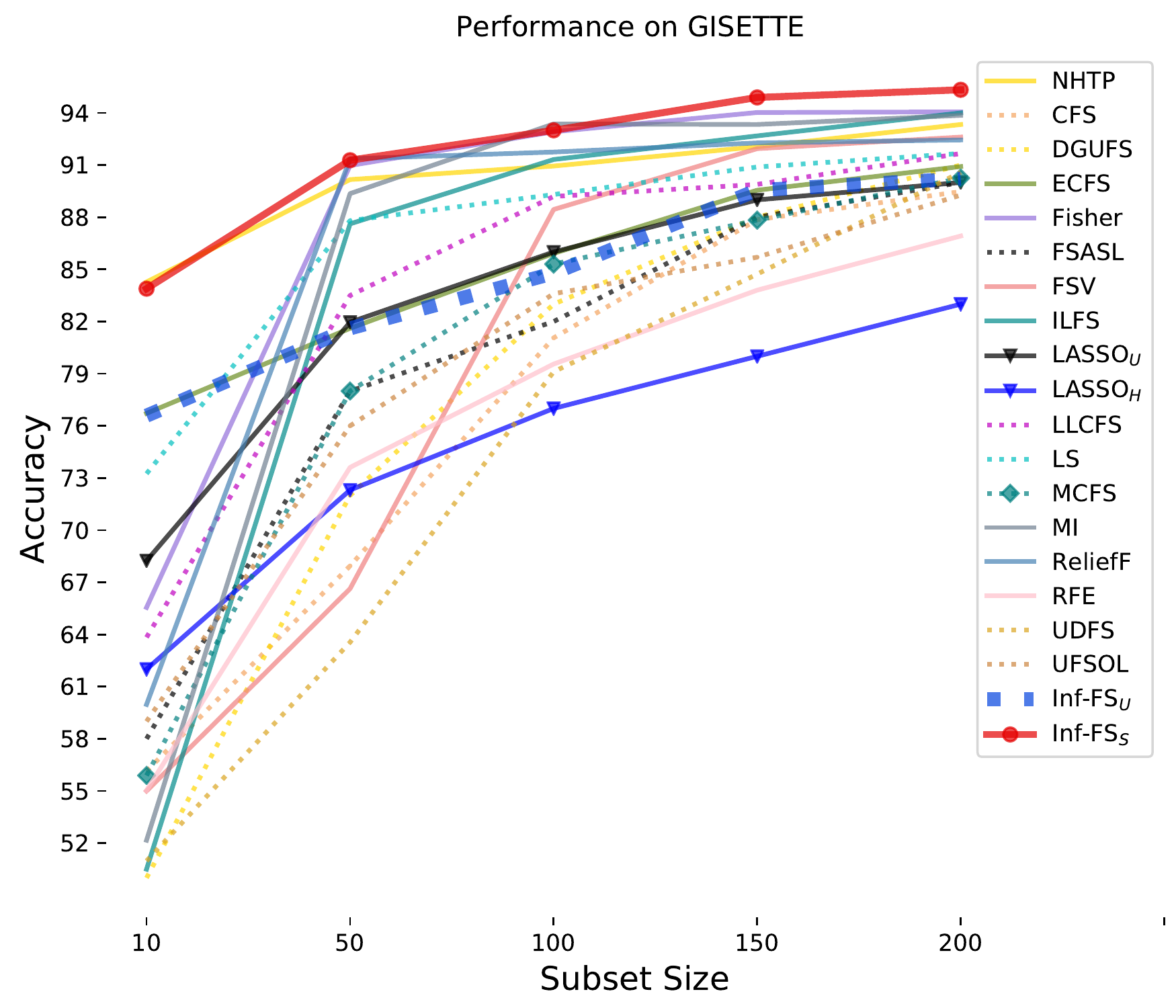}
\includegraphics[width=0.45\textwidth]{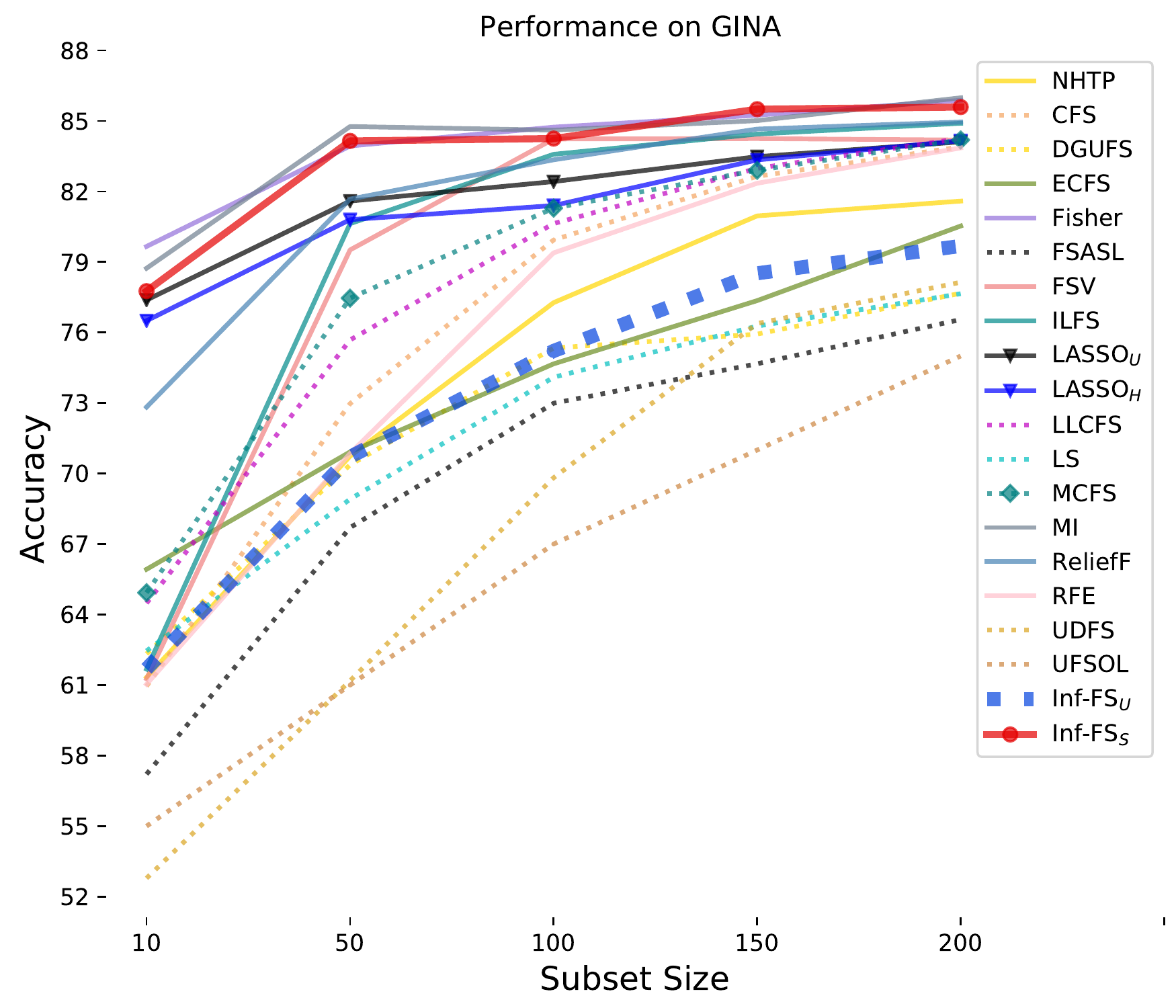}
\includegraphics[width=0.45\textwidth]{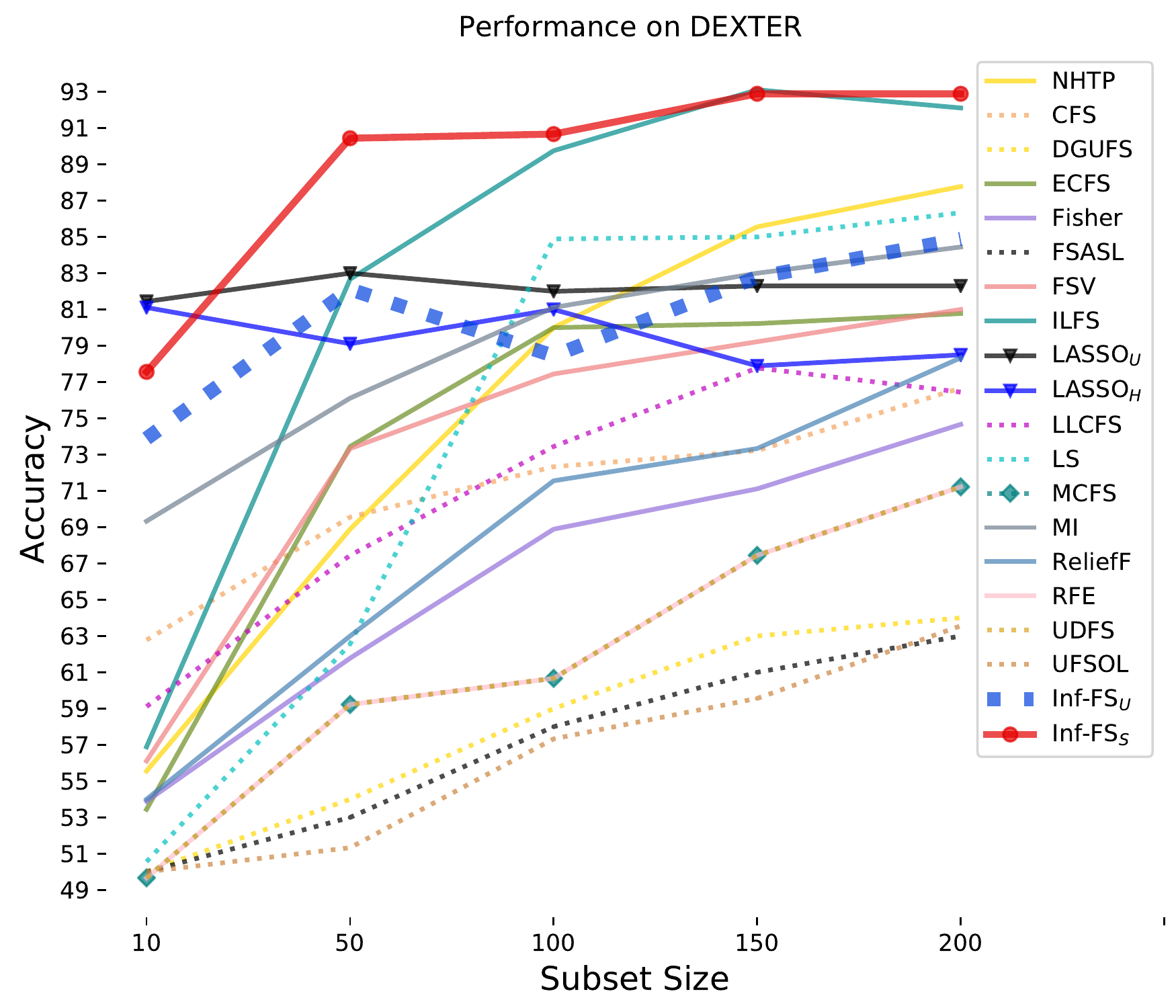}
\includegraphics[width=0.45\textwidth]{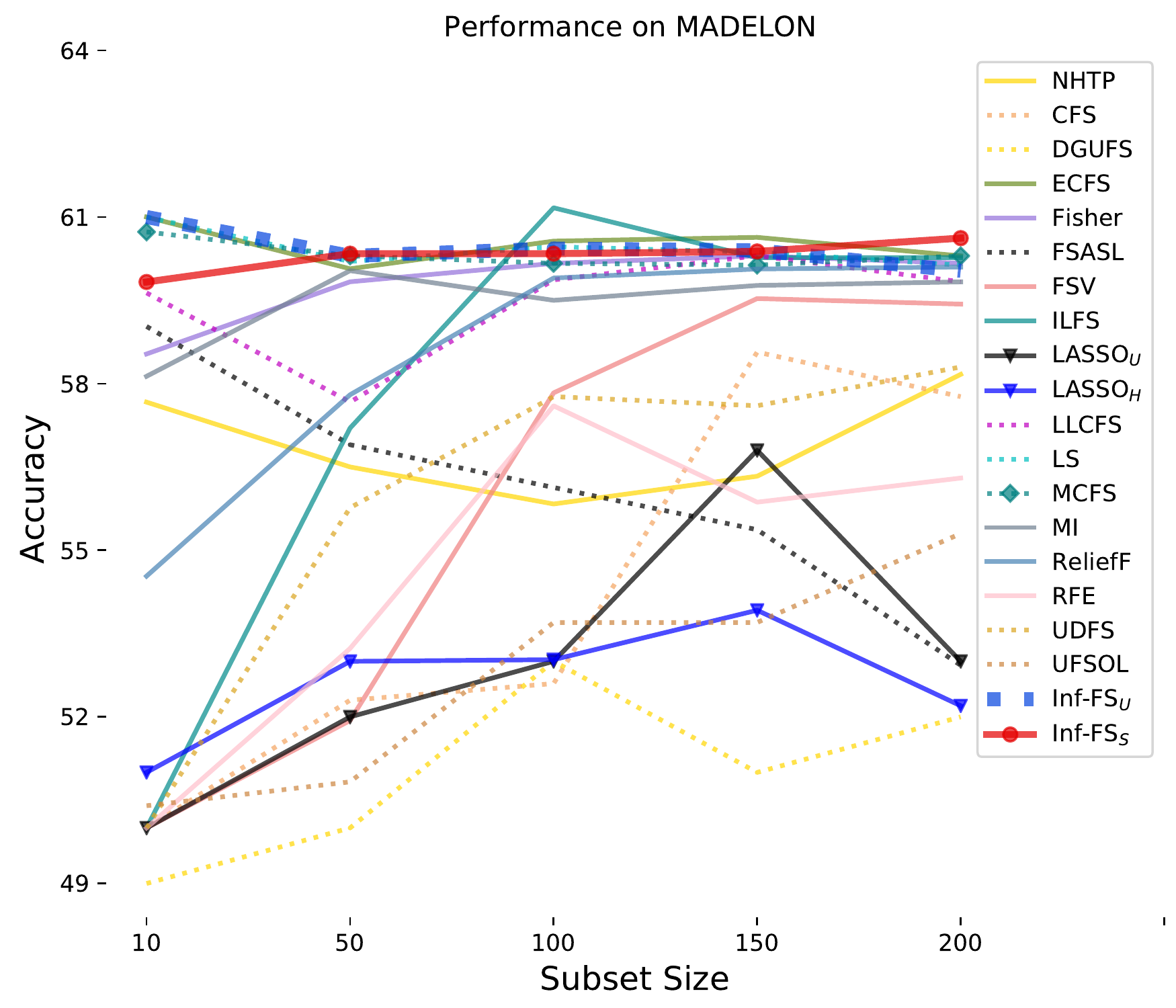}
\caption{Comparison between Inf-FS$_U$ and Inf-FS$_S$. All the supervised approaches are reported by solid lines and the unsupervised ones by dotted lines. Results are expressed in terms of classification accuracy (\%).}\label{fig:EXPSET2_NIPS}
\end{figure*}

Comparing  Inf-FS$_U$ and Inf-FS$_S$ (Fig.~\ref{fig:EXPSET1_BIO}, left and right)
one can see that, in general, Inf-FS$_S$ works better than Inf-FS$_U$, since it uses class-label information to guide the FS process. Nonetheless, it is worth knowing (no curves are reported here) that on some datasets (COLON, LEUKEMIA and LUNG) the performance of the two approaches is comparable. This interesting aspect will be further discussed in Sec.~\ref{sec:NIPS} and Sec.~\ref{sec:VOC}.

\subsection{\emph{Challenge 2}: Inf-FS$_U$ VS  Inf-FS$_S$ }\label{sec:NIPS}
This section compares the supervised and unsupervised versions of Inf-FS. Essentially, the difference between the two approaches consists of the type of functions used for weighting the graph. In fact, Inf-FS$_U$ does not employ any class-label information according to Eq. \ref{eq:unsupWeight}, while Inf-FS$_S$ is a combination of three different terms, two of them making use of the class labels (Fisher criterion and mutual information, see Eq.~\ref{eq:supWeight1}).
When the difficulty of a classification problem depends on classes that overlap, Inf-FS$_S$ can naturally favour those features that best represent the explanatory factors of the dissimilarity among the classes. 
On the other side, Inf-FS$_S$ suffers when features are severely correlated, even if they are representative for a specific class. In this case, variance and correlation computed by Inf-FS$_U$ do represent a very convenient option. 

To validate these considerations, we consider four additional datasets from the \emph{NIPS} feature selection challenge, namely:  DEXTER~\cite{guyon2004result}, GISETTE, MADELON~\cite{NIPS2003} and GINA~\cite{GINA}. GISETTE and GINA present severely overlapped classes. Indeed, the GISETTE dataset~\cite{guyon2003design} has instances of ``4'' and ``9'', two confusable handwritten digits (i.e., two overlapped classes) extracted from the MNIST data ~\cite{mnisthandwrittendigit}. Features consist of normalized pixels and quantities derived from their combination.  

The task of GINA is again handwritten digit recognition, but in this case, the two classes are \emph{even and odd} 2-digit numbers. Obviously, only the unit digit is informative. In addition to the overlapping issues among the single digits (which are taken again from the MNIST data), a further consistent overlap is caused by the digits indicating the tens. 

As for a dataset with non-descriptive features, we selected the DEXTER dataset ~\cite{guyon2004result}, composed by sparse continuous bag-of-words histograms, extracted from the Reuters text categorization benchmark~\cite{guyon2003design}. Noise is coming from $10,053$ distractors (features having no discriminative power) put voluntarily in the dataset. 

A benchmark where Inf-FS$_U$ should perform comparably if not superior to Inf-FS$_S$ is MADELON ~\cite{NIPS2003}. In fact, MADELON is an artificial dataset containing data points grouped in 32 clusters placed on the vertices of a five-dimensional hypercube and randomly labelled +1 or -1. The five dimensions constitute 5 informative features. 15 linear combinations of those features were added to form a set of 20 (redundant) informative features. Based on those 20 features one must separate the examples into the 2 classes (corresponding to the +1, -1 labels). A number of distractor features (480) called ``probes" have no predictive power. Other than this, correlated features are present. 

The results are shown in Fig.~\ref{fig:EXPSET2_NIPS}. In general, Inf-FS$_S$ outperforms Inf-FS$_U$ on DEXTER, GINA and GISETTE and achieves a absolute top performance in most of the cases. On the other hand, Inf-FS$_U$ achieves a better performance on MADELON at 10 features w.r.t. the supervised counterpart, by discarding the several correlated features in the set, and behaves comparably with Inf-FS$_S$ at the other regimes.

\rev{Considering each dataset separately, on GISETTE (Fig.~\ref{fig:EXPSET2_NIPS} top-left) Inf-FS$_S$ betters all the comparative approaches when using 10 features, having NHTP close to its performance, while in the other supervised cases the gap is substantial. Unsupervised approaches do comparably to supervised ones when it comes to 10 features, but this is probably due to the fact that 10 features are definitely too few over the 5K which are originally available, and where many of them are probably equally useful. In fact, when the number of allowed features is growing (150, 200), it is visible that most supervised approaches better the unsupervised ones.  Among the unsupervised approaches, our Inf-FS$_U$ ranks approximately third after LLCFS~\cite{zeng2011feature} and  LS~\cite{HCN05a}, since the former is driven by variance and correlation, and this does not allow to unveil features which are overlapped among classes. Notably, LLCFS~\cite{zeng2011feature} and  LS~\cite{HCN05a} select features which are locality preserving, i.e., which agree on a clustering over the data. We may think that this clustering is capable to naturally separating the digits data, providing a more powerful solution than Inf-FS$_U$.}        

\rev{On GINA instead  (Fig.~\ref{fig:EXPSET2_NIPS} top-right), supervised approaches show at just 10 features a consistent advantage over the unsupervised methods. Here, Inf-FS$_S$ is on pair with the mutual information MI~\cite{Hutter:02feature} and the Fisher approach~\cite{Quanquanjournals}. 
In fact, Inf-FS$_S$ contains both of them in the adjacency matrix $A$ (see Sec.~\ref{sec:supA}), and they are useful to highlight features that do not overlap across classes, i.e., which are non linearly correlated with the class information. Inf-FS$_U$ gives here the worst performances, ranking approximately fourth with respect to slower and more complex approaches (MCFS~\cite{Cai:2010}, LLCFS~\cite{zeng2011feature}, DGUFS~\cite{JunGuo_AAAI_2018_DGUFS}) which once again exploit the hypothesis that data is organized in multiple clusters which we are ignoring with Inf-FS$_U$.}

\rev{DEXTER (Fig.~\ref{fig:EXPSET2_NIPS} bottom-left) has the highest number of features (20K) so that restricting to only 10-200 features opens to many equivalent selections, which anyway are better individuated by INf-FS$_S$ (among the supervised approaches, except the 10 features case where LASSO shows to be better) and by INf-FS$_U$ (among the unsupervised approaches, on pair with LS~\cite{HCN05a} which is better at 100-200 features).}

\rev{On MADELON we already have discussed above the results of Fig.~\ref{fig:EXPSET2_NIPS} (bottom-right) .}

\subsection{\emph{Challenge 3}: Feature selection on CNN Features}\label{sec:VOC}
 \begin{figure*}[!]
\centering
\includegraphics[width=0.48\textwidth]{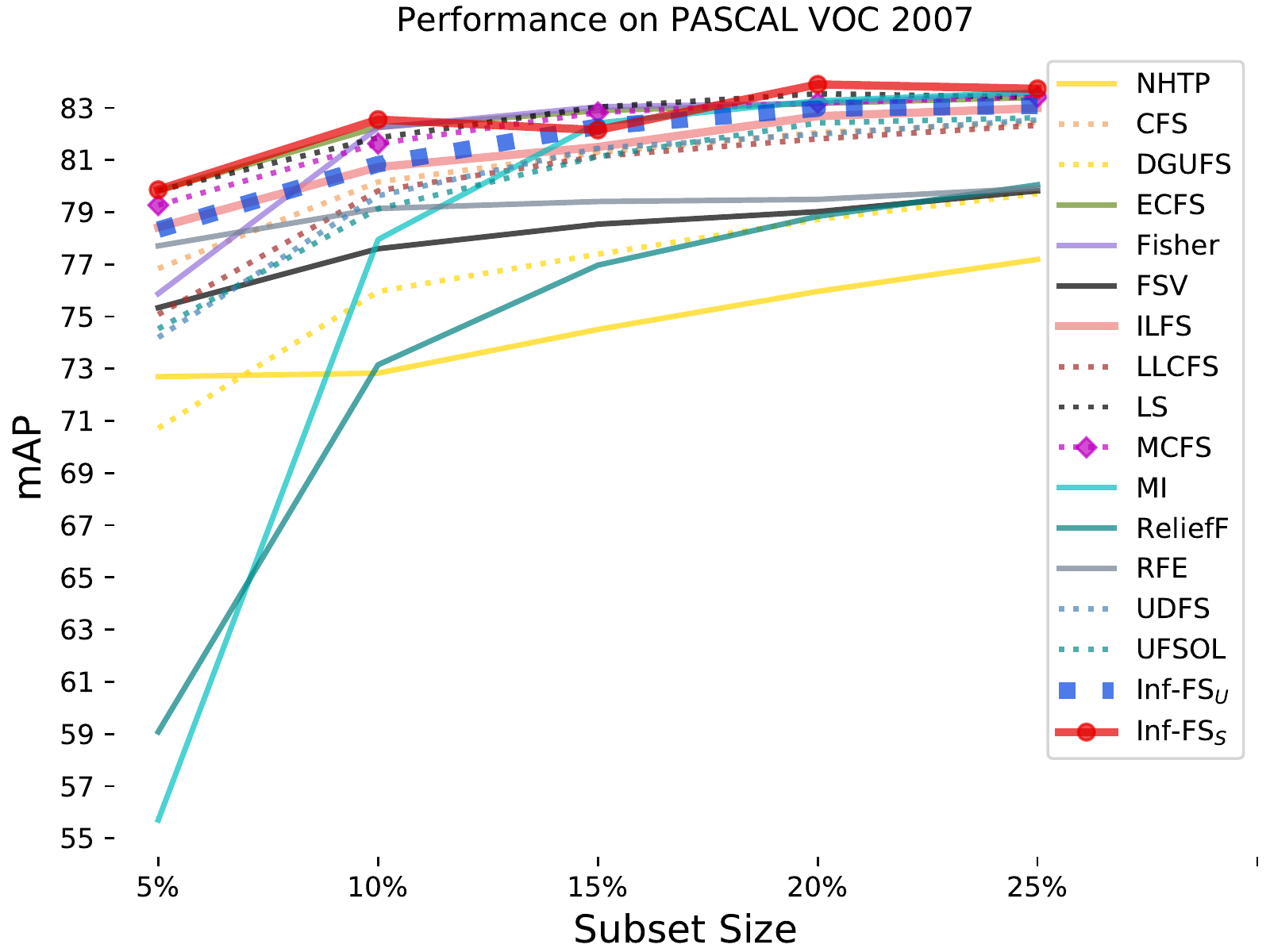}
\includegraphics[width=0.48\textwidth]{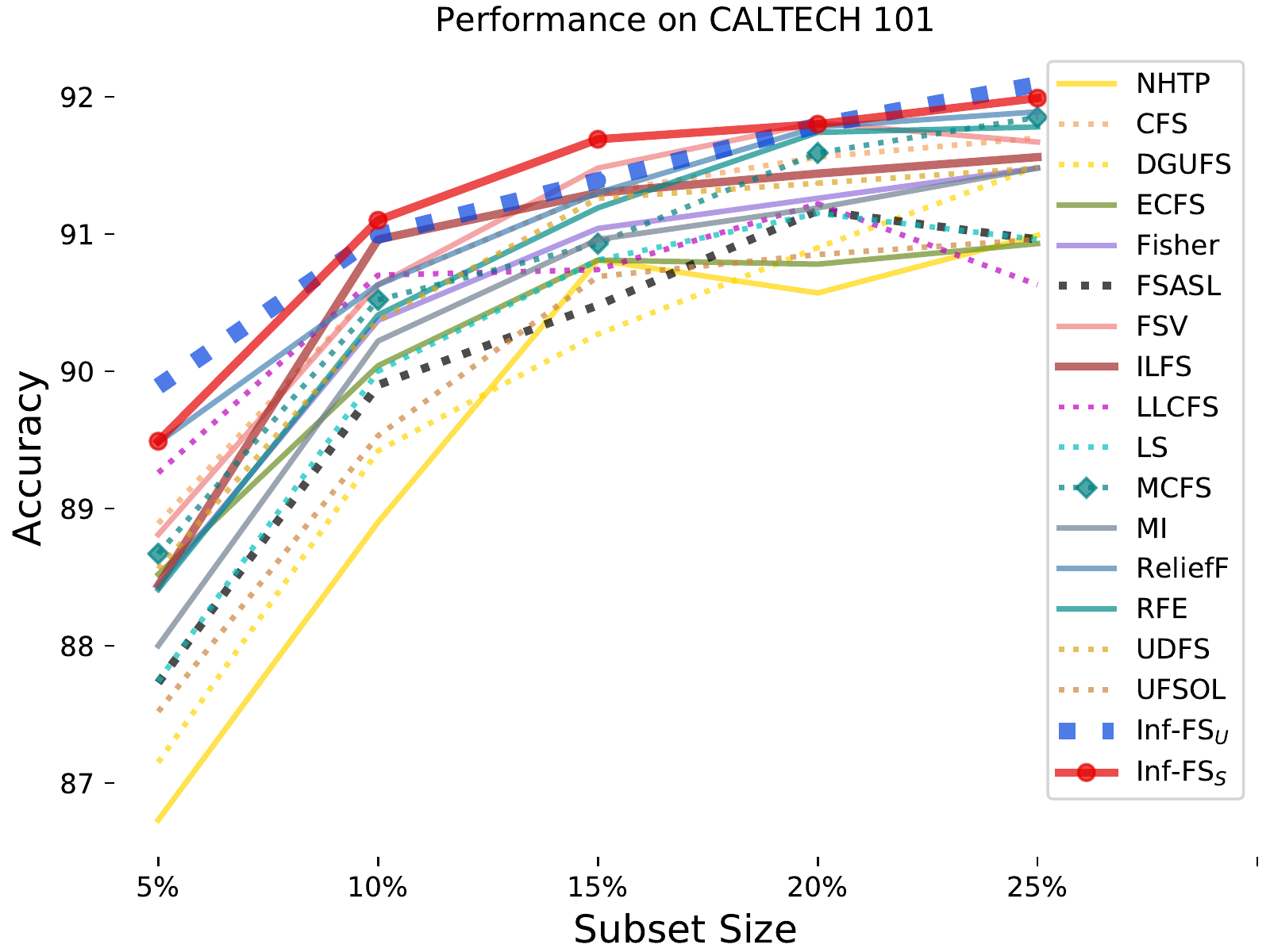}
\caption{Performance achieved for the image classification task reported in terms of mAP (VOC 2007) and classification accuracy (Caltech-101) while selecting the first 5\%, 10\%, 15\%, 20\%, and 25\% features. Solid lines individuate supervised feature selection approaches, dotted lines indicate unsupervised approaches.}\label{fig:EXPSET3_VOT}
\end{figure*}

\begin{figure*}[!]
\centering
\includegraphics[width=0.49\textwidth]{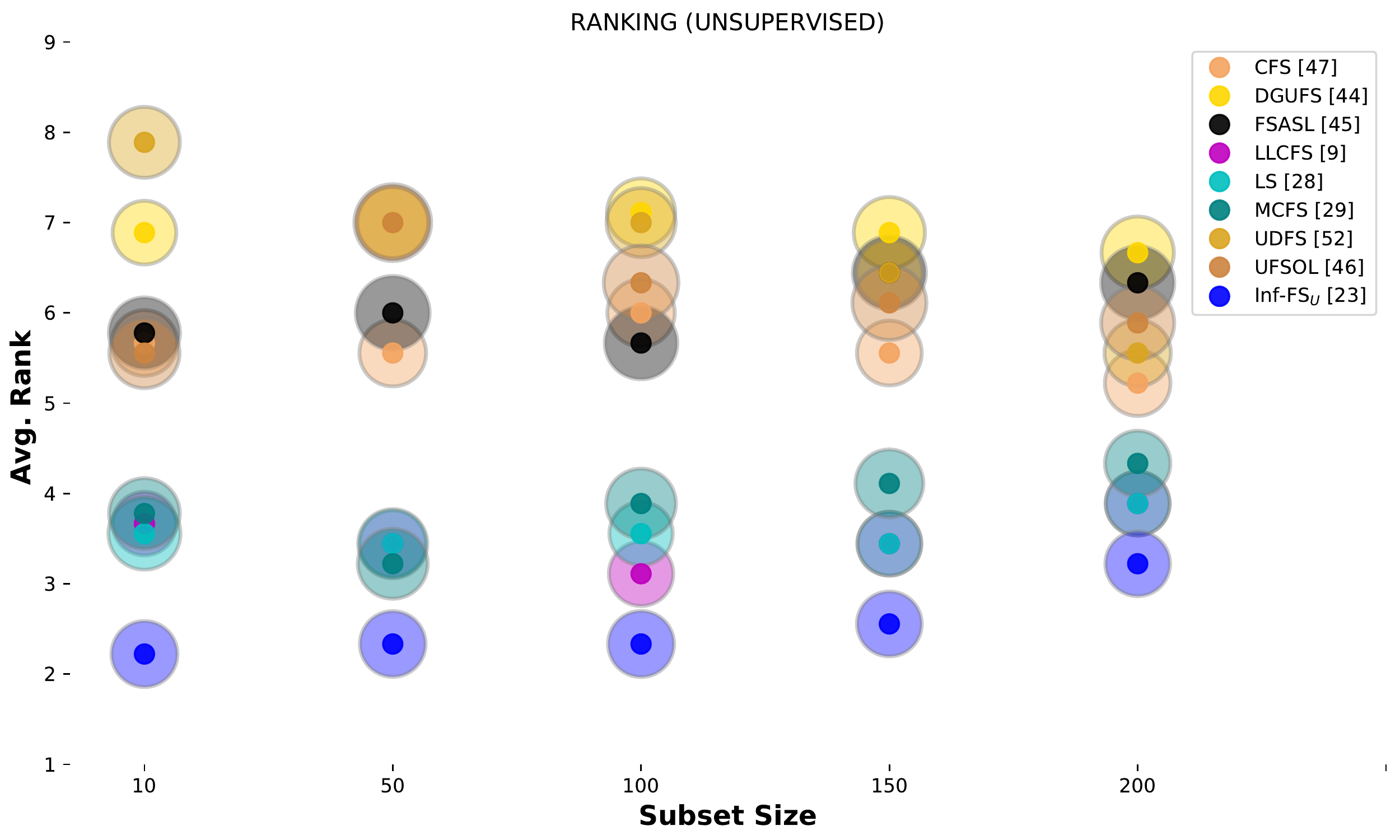}
\includegraphics[width=0.49\textwidth]{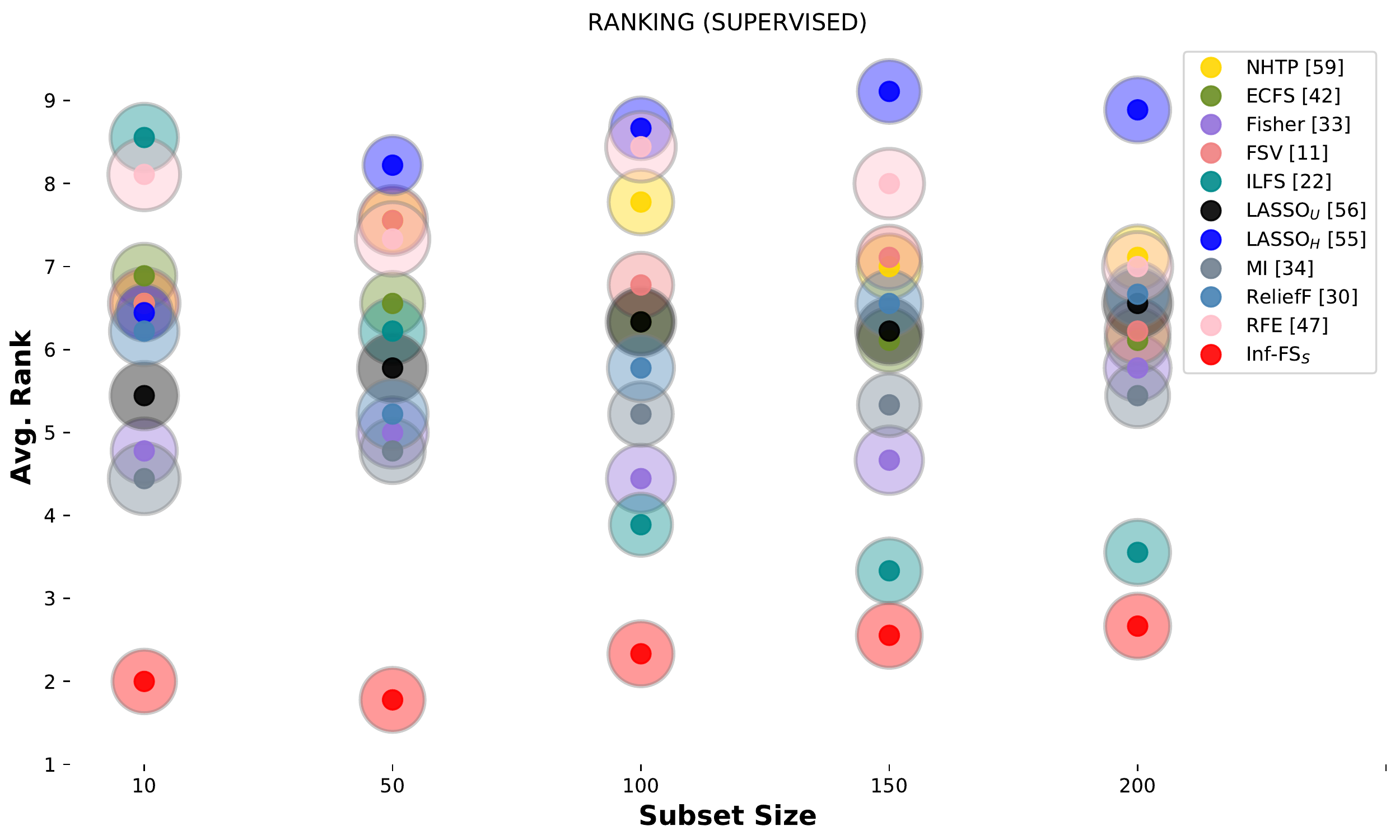}
\caption{Bubble plot showing the average ranking performance (y-axis) overall the datasets while increasing the number of selected features for the unsupervised approaches (left) and supervised ones (right). The area of each circle is proportional to the variance of the ranking. }\label{fig:ALL}
\end{figure*}

Applying feature selection on deep learning-based cues is a recent trend in image recognition~\cite{denil2013predicting,yu2018_NISP}. In fact, recent studies show that feature learning and deep learning are not immune to produce redundant or introduce useless information in the learned representations. For example, \cite{yu2018_NISP} proposed a generic framework for network compression and acceleration where CNNs are pruned by removing neurons with least importance, resulting in more robust networks. Neuron importance scores (usually associated to the last layer of the network, before classification) are computed by Inf-FS$_U$ as a function of the importance of all the other neurons in the layer. 
 
In this subsection, we evaluate the performance of the proposed approach on features learned by the very deep ConvNet~\cite{Simonyan14c} framework, where the pre-trained model used for the ImageNet Large-Scale Visual Recognition Challenge 2014 (ILSVRC) is adopted. We use the $4,096$-dimension activations of the last layer as image descriptors (L2-normalized afterwards), and we focus on the CALTECH 101 and PASCAL VOC-2007 datasets. These datasets allow for a systematic testing of the feature selection approaches taken into account in this paper, in a reasonable amount of time. We omit to choose other benchmarks (Imagenet for example) since for some of the comparative methods (LASSO and MCFS) the running time for a single trial is exceeding the week. Indeed, for each comparative approach, we perform a total of $200$ runs. 

According to the experimental protocol provided by the VOC challenge, a one-vs-rest SVM classifier is trained for each class (where cross-validation is used to find the best parameter C) and evaluated independently.
Fig.~\ref{fig:EXPSET3_VOT} reports the performance curves obtained with the 18 feature selection approaches (solid lines for supervised approaches, dotted lines for unsupervised ones). In this case, the goal was to investigate the classification while keeping the first 5\%, 10\%, 15\%, 20\%, 25\% of the features, corresponding to 205, 410, 614, 819 and 1024 characteristics. 

From Fig.~\ref{fig:EXPSET3_VOT} (Left), it can be seen that the supervised Inf-$FS_S$ reaches good performance in general, with a slightly superior performance w.r.t. the eigenvector centrality-based approach (ECFS). In general, the supervised approaches are organized into two groups, the most performing ones are the INFFS, ECFS, that, together with MI and ILFS gives an increase in the classification performance when adding more features. The other supervised approaches (RFE, FSV and RELIEF) seems to have a lower trend. Viceversa, all of the unsupervised approaches are more consistent among themselves, with Inf-FS positioning in the top 3 positions after LS and LLCFS. In the case of CALTECH 101 it is easy to see that the task is easier, with all of the approaches positioning in a narrow band of performance. Notably, Inf-FS$_S$ and Inf-FS$_U$ are on pair at the top position. 

\begin{figure}[!]
\centering
\includegraphics[width=0.50\textwidth]{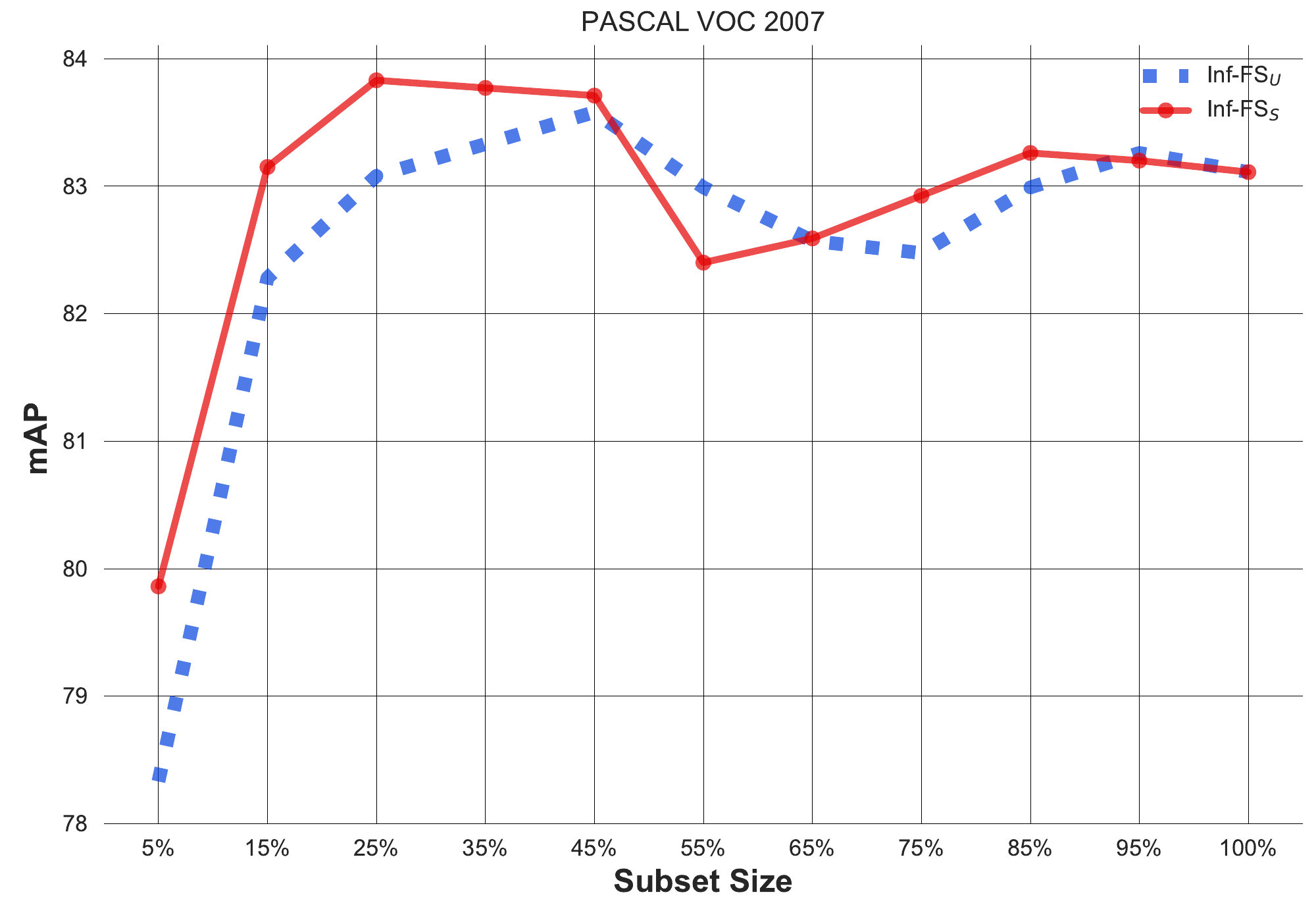}
\caption{Varying the cardinality of the selected features on VOC 2007. Mean average precision instead of classification accuracy is provided here.}\label{fig:EXPSET3_full}
\end{figure}

On the PASCAL 2007, we performed an additional experiment, aimed at exploring the performances when spanning the number of features retained from 5\% to 100\% (Fig.~\ref{fig:EXPSET3_full}). The idea is to check how much difference holds when keeping a small number of features with respect to the whole set. In fact, feature selection approaches often represent a compromise between admitting a lower classification performance at the price of a faster time of task execution~\cite{Guyon:2003:IVF:944919.944968}. We apply both Inf-FS$_S$ and Inf-FS$_U$. Noteworthy, both of the approaches provide features subsets leading to a performance (mAP) superior to the one obtained with the entire pool. In particular, with 25\% of features, Inf-FS$_S$ raises the classification performances of barely 1 percentage point (83.8\% against 83.1\% fo the full set). Better performances are obtained in the range of 25\%-45\%. The Inf-FS$_S$ shows that there is a 10\% of features ranked last which cause a slight bending of the performances (see the 90\%-100\% range). Inf-FS$_U$ has a similar behaviour, but lower in mAP score: The peak is at 45\% of features (83.6\%). To further explore the behavior of the approach in the range of best performance (25\%-45\% for Inf-FS$_S$ and 35\%-45\% for Inf-FS$_U$) we perform a fine-grained cardinality analysis reaching the absolute best of Inf-FS$_S$ at 31.5\% features (84.18\% mAP) and 36.5\% for Inf-FS$_U$ (83.91\% mAP).

 \begin{table*}[!h]
\small
\centering
\resizebox{0.9\textwidth}{!}{%
\begin{tabular}{| l | c c | c c |  c c | c c |}
\cline{1-9}
\textbf{Dataset}  & \multicolumn{2}{|c|}{\textbf{LASSO (unhinged) ~\cite{van2015learning}}} & \multicolumn{2}{|c|}{\rev{\textbf{LASSO (hinge)}~\cite{van2008high}}} & \multicolumn{2}{|c|}{\textbf{Inf-FS$_U$}}& \multicolumn{2}{|c|}{\textbf{Inf-FS$_S$}} \\
\cline{1-9}
\hline 
      &     Subset             &   Best Prev. Perf. &     Subset             &   Best Prev. Perf. &   Subset             &   Best Prev. Perf.  &  Subset  & Best Prev. Perf. \\\hline
DEXTER& (10) 80.3\% & (50) 82.9\% & (2343)  79.9\% & (10)  81.1\% &  (466) 83.8\%&(200) 84.8\% & (339) \textbf{ 92.8\%} & (150) 92.9\% \\ 
GISETTE & (2126) \textbf{ 95.3\%} & (200) 90.3\% &(2482)  85.9\% & (200)  83.5\% &   (707) 87.7\% &(200) 90.2\% & (638) 94.1\% & (200) 93.3 \\ 
GINA & (478) \textbf{ 83.8\% }& (200) 84.1\% & (485)  80.4\% & (200)  84.2\% &   (152) 76.2\% & (200) 79.6\% &(127) 83.4\% & (200) 85.6\% \\ 
MADELON &(233)  55.9\%& (150) 56.9\% & (396)  54.3\% & (150)  53.9\% &   (48)  \textbf{ 58.7\% }&(10) 61.0\%& (32) 57.1\%  & (200) 60.0  \\ 
\hline 
\emph{COLON} &(22) 66.7\% &(50) 85.5\% & (1131)  84.4\% & (200)  80.0\% &    (326) \textbf{91.1\%} & (150) 92.7\% & (174) \textbf{91.1\%}   & (100) 92.7\%  \\
\emph{LEUKEMIA} & (18)  79.5\% &(200) 97.1\% & (1810)  93.8\% & (150)  93.3\% &   (618) 94.7\% & (10) 94.7\% & (242) \textbf{95.2\%}  & (10) 94.8\%  \\
\emph{PROSTATE} &(43)  87.0\% &(100) 95.3\% & (3168)  90.7\% & (150)  93.7\% &    (1014) 93.0\% & (100) 93.3\% &(563) \textbf{94.7.6\%}  & (150) 96.6\% \\
\emph{LYMPHOMA} &(13) 56.7\% &(150) 91.6\% & (2105)  86.7\% & (150)  75.8\% &    (674)  93.3\% & (150) 93.3\% & (395) \textbf{95.8\%} & (200) 98.3\%  \\
\emph{LUNG} & (49)  89.8\% &(50) 96.6\% & (5297)  96.2\% & (150)  97.6\% &   (400) \textbf{ 99.9\%}  & (200) 99.8\% & (361) \textbf{ 99.9\%}  & (200) 99.8\%  \\
\hline \vspace{0.02cm}
VOC 2007  & N/A & N/A & N/A & N/A & (1,883)   83.6\% & (1024) 83.1\% & (696)   \textbf{ 83.5\%}  & (819) 83.8\% \\
CalTech 101  & N/A & N/A & N/A & N/A & (2250)  \textbf{  92.0\%} & (1024) 92.1\% & (942)   91.8\%  & (1024) 91.9\%\\
\hline
\end{tabular}
}
\caption{The feature subset selection results reported in terms of accuracy (\%). The values enclosed in round brackets show the number of the features kept. In bold the best performance for the \emph{Subset} selection problem.}
\label{table:subsetres}
\end{table*}

\subsection{The versatility of Inf-FS$_U$ and Inf-FS$_S$ }\label{sec:NIPS}
\rev{In this section we want to summarize the diverse experiments carried out so far, demonstrating that one of the most valuable merit of the Inf-FS framework is that \emph{it applies favorably on every genre of feature selection scenario}.  To this sake, we set up in Fig.~\ref{fig:ALL}
two bubble-plots showing the average ranking (the lower, the better) for each compared approach (y-axis), considering all of the used datasets (except CALTECH 101 and PASCAL VOC where LASSO did not apply, and where we evaluated different numbers of features), separating the unsupervised and supervised approaches that we have considered in the experiments. }

\rev{In practice, the ranking represents the position of an approach (as classification accuracy) with respect to all the others. In the case a given approach has the best accuracy for a given benchmark, its rank on that benchmark is 1, in the case it gives the second-best accuracy the rank is 2, and so on. The average ranking shows how an approach, independently on the accuracy score, is \emph{generically better} than the others, exhibiting a relative ordering.}

\rev{The average ranking is computed with respect to different subsets of features (x-axis), and is enriched by the standard deviation  in the ranking (how consistently an approach had a particular rank), depicted by the size of the blob (the larger the size, the higher the ranking variance).}

\rev{The figures convey a clear message, since both Inf-FS unsupervised and supervised have the best rank, with a variance of 0.23 which indicates a stable behavior of both the approaches. Notably, Inf-FS$_S$ is definitely the most effective choice when it comes to few features selected; the mutual information-based MI~\cite{Hutter:02feature} and the Fisher criterion for feature selection~\cite{Quanquanjournals} follow.  In the case of unsupervised approaches, Inf-FS$_U$ is first, followed by the clustering based approaches  LS~\cite{HCN05a} and MCFS~\cite{Cai:2010}.}

\subsection{\emph{Challenge 4}: Automatic Subset Selection}\label{sec:SUB}
In this section, we test the process of selecting a subset of relevant features from the ranking provided by Inf-FS, explained in  Sec.~\ref{sec:feature selection}. 

To this sake, we repeat all of the experiments with 
Inf-FS$_S$ and Inf-FS$_U$ on the 11 datasets examined so far, selecting as relevant features the ones indicated by the cluster which includes the first-ranked feature, and using them for the classification tasks. As comparative approach, we consider LASSO learned with hinge loss~\cite{van2008high} and unhinged loss~\cite{van2015learning}, since it is the only which allows to automatically select a precise number of features, that is, the ones which survive the shrinking process during the training stage. \rev{In particular, we individuate the best-performing LASSO by 5-fold cross-validating the regularization parameter over the training set of each benchmark, for both the hinged and unhinged versions.} The results are reported in Table~\ref{table:subsetres}

For each pair $<dataset, method>$, we report four different quantities: in the \emph{Subset} column we show in round brackets the number of selected features, and alongside the classification accuracy obtained with that number of features. In the \emph{Best Prev. Perf.} column, we report in round brackets the number of features that provided the best performance obtained \emph{in the previous experiments} (following on the right). In the table, bold scores indicate the highest classification performance among the scores obtained by the automatic selection of feature subset, not the highest absolute.

From the results, several observations can be drawn: 
\begin{itemize}
\item \rev{The automatic selection of the number of features allow Inf-FS$_U$ and Inf-FS$_S$ to provide higher performances than LASSO on 9 out of 11 cases, with LASSO unhinged beating the Inf-FS framework on GISETTE and GINA;}
\item Tightly connected with the previous point, and worth noting, the Inf-FS framework selects definitely less features than the LASSO  approaches (apart from the microarray datasets, where anyway LASSO unhinged is giving scarce performance). LASSO unhinged tends to keep features in a number which is highly variable; for example, it suggests a very large amount of features (2126 for GISETTE) or very few (the five microarray datasets); this seems to be correlated with the number of samples in the dataset, that, for the microarray datasets, is quite small. LASSO hinge appears to be more stable (but it gives the highest number of features).
\item Inf-FS$_S$ requires for all of the datasets less features  than Inf-FS$_U$ (operating with the automatic selection), showing that the class information enriches the discriminative power of the cues. 
\item Inf-FS performance with the automatic selection remain competitive in every scenario, while LASSO unhinged performs very poor on the small-sample, high dimensional case. 
\end{itemize}

\section{Conclusions}\label{sec:conc}
\rev{In this work we considered the feature selection problem under a brand-new perspective, i.e., as a regularization problem, where features are nodes in a weighted fully-connected graph, and a selection of $l$ features is a path of length $l$ through the nodes of the graph.
Under this view, the proposed Inf-FS framework associates each feature to a score originating from pairwise functions (the weights of the edges) that measure relevance and non redundancy. This score has different explanations: under a power series of matrices view indicates the value that a feature can bring in a possibly infinite selection of features.  Alternatively, under an absorbing Markov chain perspective, the score indicates how many times a feature would be associated to the other cues as complementary, before to end the process of selection. A precise subset of features can be provided, by examining the distribution of these scores.}

\rev{Inf-FS can be customized by hand-crafting the pairwise functions, and here we presented two customizations, for unsupervised and supervised scenarios, respectively. Future work will be spent in designing an end-to-end system capable to infer the optimal pairwise functions.  
}

\ifCLASSOPTIONcompsoc
  \section*{Acknowledgments}
\else
  \section*{Acknowledgment}
\fi

This work is partially supported by the Engineering and Physical Sciences Research Council (EPSRC) under grant EP/N035305/1, and by the project of
the Italian Ministry of Education, Universities and Research (MIUR) "Dipartimenti di Eccellenza 2018-2022".

\ifCLASSOPTIONcaptionsoff
  \newpage
\fi



\bibliographystyle{IEEEtran}
\bibliography{egbib}
%



%

\begin{IEEEbiography}[{\includegraphics[width=1in,height=1.25in,clip,keepaspectratio]{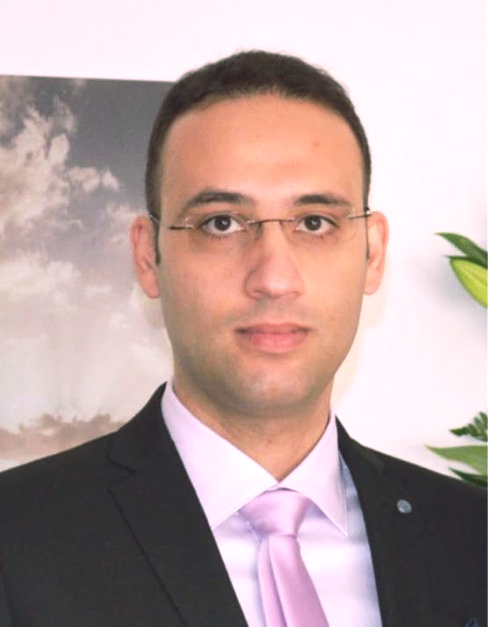}}]{Giorgio Roffo} is 
with the University of Glasgow where he is a Research Associate at the School of Computing Science. He received the European PhD degree in computer science from the University of Verona, Italy. Previously, he was with the Italian Institute of Technology (IIT), Genoa, Italy. His primary research interests are in the areas of machine learning, deep learning and computer vision. He contributed to the research field by publishing more than $10$ articles in prestigious journals and conferences. He is in the technical program committee of leading conferences in computer vision and pattern recognition.
\end{IEEEbiography}

\begin{IEEEbiography}[{\includegraphics[width=1in,height=1.25in,clip,keepaspectratio]{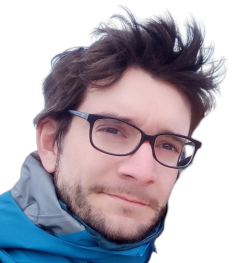}}]{Simone Melzi}
is a Post Doctoral researcher at Università degli Studi di Verona (Italy). He received his PhD in Computer Science at Università degli Studi di Verona (2018) and graduated in math summa cum laude from the University of Milan ``La Statale" (2013). He received the EG-Italy PhD thesis award (2018). His main research interests are geometry processing, shape matching and 3D shape analysis. He has authored over 10 publications in leading journals and conferences.
\end{IEEEbiography}

\begin{IEEEbiography}[{\includegraphics[width=1in,height=1.25in,clip,keepaspectratio]{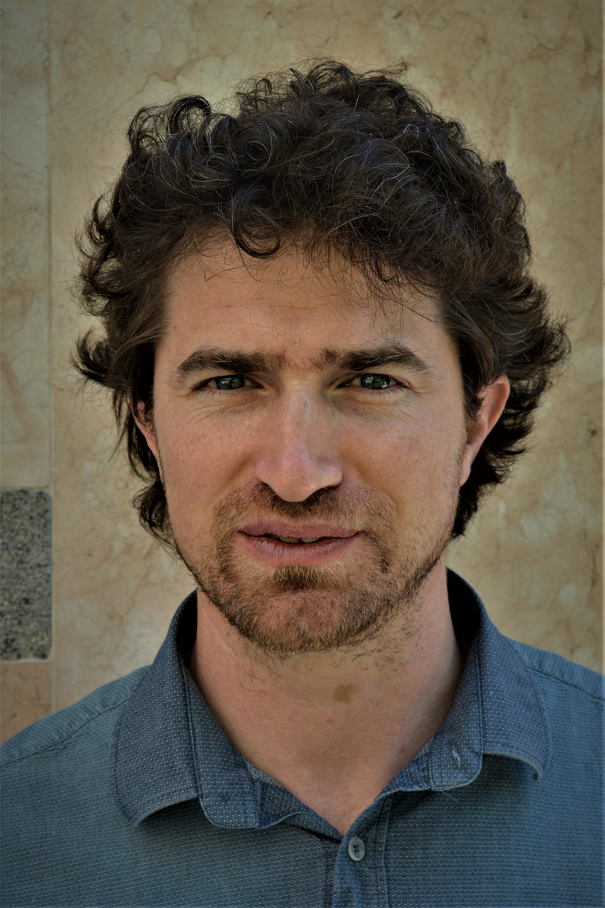}}]{Umberto Castellani}
is Associate Professor of the Department of Computer Science at University of Verona. He received his Dottorato di Ricerca (PhD) in Computer Science from the University of Verona in 2003 working on 3D data modelling and reconstruction. He held visiting research positions at Edinburgh University (UK), Universite' Blaise Pascal (France), Michigan State University (USA), Universite' D'Auvergne (France), Italian Institute of Technolgy (IIT), and University College London (UK). His research is focused on 3D geometry processing, statistical learning and medical image analysis. He is working on 3D shape processing from several acquisition systems for modelling, analysis and recognition. He has coauthored  more than 130 papers published in leading conference proceedings and journals. He is member of the editorial board of the Pattern Recognition journal and member of Program Commettee of several workshops and conferences. His research activity is mainly developed within European (EU) projects, national (MIUR) projects, and projects funded by private companies.
\end{IEEEbiography}

\begin{IEEEbiography}
[{\includegraphics[width=1in,height=1.25in,clip,keepaspectratio]{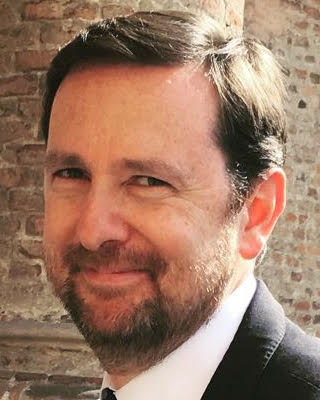}}]
{Alessandro Vinciarelli} (\textit{http://vinciarelli.net}) is with the University of Glasgow where he is Full Professor 
at the School of Computing Science and Associate Academic at the Institute of Neuroscience and Psychology. 
His main research interest is in Social Signal Processing, the domain aimed at modeling analysis and synthesis 
of nonverbal behavior in social interactions. Overall, he has published more than 150 works, including one authored 
book, and 35 journal papers. He has been General Chair of the IEEE International Conference on Social Computing 
in 2012 and of the ACM International Conference on Multimodal Interaction in 2017. He is or has been Principal 
Investigator of several national and international projects, including a Centre for Doctoral training (texttt{http://social-cdt.org}), 
a European Network of Excellence (the SSPNet, \textit{www.sspnet.eu}), and more than 10 projects funded by the 
Swiss National Science Foundation end the UK Engineering and Physical Sciences Research Council. Last, but not
least, Alessandro is co-founder of Klewel (\textit{www.klewel.com}), a knowledge management company recognized with 
national and international awards, and scientific advisor of Neurodata Lab (\textit{http://neurodatalab.com}).
\end{IEEEbiography}
\begin{IEEEbiography}[{\includegraphics[width=1in,height=1.25in,clip,keepaspectratio]{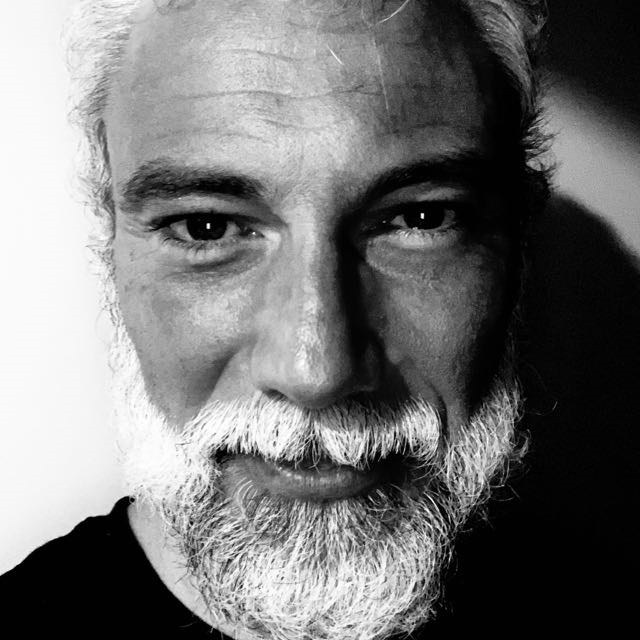}}]{Marco 
Cristani} is Associate Professor (Professore Associato) at the Computer 
Science Department, University of Verona, Associate Member at the National Research Council (CNR), External Collaborator at the Italian Institute of Technology (IIT). His main research interests are in statistical pattern recognition and computer vision, mainly in deep learning and generative modeling, with application to social signal processing and fashion modeling. On these topics he has published more than 170 papers, including  two edited volumes, 6 book chapters, 40 journal articles and 129
conference papers. He has organized 11 international workshops, co-founded a spin-off company, Humatics, dealing with e-commerce for fashion.  He is or has been Principal Investigator of several national and international projects, including PRIN and H2020 projects. He is member of the editorial board of the Pattern Recognition  and Pattern Recognition Letters journals. He is Managing Director of the Computer Science Park, a technology transfer centre at the University of Verona. Finally, he is ACM, IEEE and IAPR member.

\end{IEEEbiography}




\end{document}